\documentclass[10pt,twocolumn,letterpaper]{article}

\usepackage{cvpr}
\usepackage{times}
\usepackage{epsfig}
\usepackage{graphicx}
\usepackage{amsmath}
\usepackage{amssymb}
\usepackage{bbm}

\usepackage{xcolor}
\usepackage{booktabs}
\usepackage{multirow}
\usepackage{multicol}
\usepackage[flushleft]{threeparttable}
\usepackage{caption}
\usepackage{subfig}
\usepackage{stackengine}
\usepackage{makecell}
\usepackage[square,sort,comma,numbers]{natbib}

\usepackage[pagebackref=true,breaklinks=true,letterpaper=true,colorlinks,bookmarks=false]{hyperref}

\cvprfinalcopy 

\def\cvprPaperID{ } 

\ifcvprfinal\fi
\begin{document}

\title{Learning Generalizable Features Across Domains \\by Disentangling Representations}

\author{Qingjie Meng\\ Imperial College London\\\and Daniel Rueckert\\  Imperial College London\\\and Bernhard Kainz\\
Imperial College London\\
}

\maketitle

\begin{abstract}
Deep learning models exhibit limited generalizability across different domains.
Specifically, transferring knowledge from available entangled domain features (source/target domain) and categorical features to new unseen categorical features in a target domain is an interesting and difficult problem that is rarely discussed in the current literature. 
This problem is essential for many real-world applications such as improving diagnostic classification or prediction in medical imaging.
To address this problem, we propose Mutual-Information-based Disentangled Neural Networks (MIDNet) to extract generalizable features that enable transferring knowledge to unseen categorical features in target domains. 
The proposed MIDNet is developed as a semi-supervised learning paradigm to alleviate the dependency on labeled data. This is important for practical applications 
where data annotation requires rare expertise as well as intense time and labor. 
We demonstrate our method on handwritten digits datasets and a fetal ultrasound dataset for image classification tasks. Experiments show that our method outperforms the state-of-the-art and achieve expected performance with sparsely labeled data.
\end{abstract}

\section{Introduction}
Deployment of deep neural networks (DNNs) in real-world scenarios is challenging because of feature distribution differences between training data and test data. This difference is known as \emph{domain shift}~\cite{joaquin2009}, which leads to poor transfer of network performance from training to testing. Domain shift is ubiquitous in many practical applications such as image classification~\cite{Saenko2010,Long2015,Tzeng2017} and image segmentation~\cite{Zhang2017,Zou2018,Dou2019,Chartsias2019}.
The problem of domain shift can be categorized into (a) covariate shift (different distributions in latent features), (b) prior probability shift (change of labels) and (c) concept shift (different relationship between latent features and the desired label)~\cite{Saito2018,Lee2019}.
Covariate shift is the key reason for the lack of generalizability of models and can result from trivial sources such as different image acquisition devices (e.g. medical imaging~\cite{Kamnitsas2016,Chen2019,Chartsias2019}), noise patterns or different combination of specific image features (e.g. shapes and textures~\cite{geirhos2018imagenet}).

\begin{figure}[t]
 \centering
 \includegraphics[width=\textwidth, trim=3.3cm 12.6cm 0cm 2cm, clip]{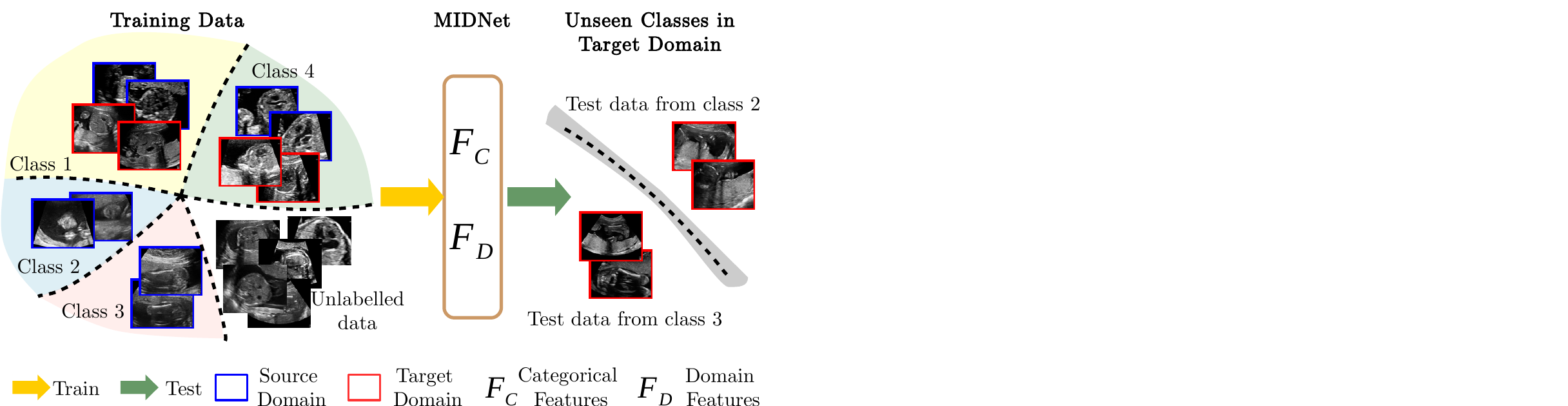}
 \caption{The proposed method (MIDNet) learns to extract generalized features across domains from sparsely labeled training data. Generalized features ($\mathcal{F}_C$, $\mathcal{F}_D$) are able to correctly classify images with previously unseen entangled categorical features and domain features. This is of particular importance for real-world applications such as improving diagnostic classification in medical imaging.}
 \label{flowchart}
\end{figure}

In contrast to the human visual system, DNNs exhibit weak generalizability when confronted with
previously unseen entangled image features.
This is the main problem we focus on in this paper. As outlined in Fig.~\ref{flowchart}, we want to improve DNN performance on unseen categories in a target domain where all categories from a source domain and a subset of categories from a target domain are available for training.
Fine-tuning DNNs on task-specific datasets is a possible solution but often infeasible due to the lack of sufficient annotated data in the target domain.
Domain adaptation algorithms have been widely studied to tackle the domain shift problem by extracting domain-invariant features, aiming to transfer knowledge from a source domain to a target domain~\cite{Peng2019}. Previous work proposed various methods from three main groups: (1) discrepancy measurement approaches, (2) adversarial adaptation approaches, and (3) generative-model-based approaches. 
The approaches in the first group aim to align the feature distributions of source and target domain by measuring the discrepancy between representations, such as Maximum Mean Discrepancy~\cite{Tzeng2014,Long2015} or correlation distance~\cite{Sun2016}. 
Instead of computing the discrepancy metric, adversarial adaptation approaches use DNNs to approximate the discrepancy of representations, which encourages the extracted features to be invariant for domain discrimination, such as transfer learning methods~\cite{Cao2019,ChenXinyang2019} and adversarial training methods~\cite{Ganin2016,Kamnitsas2016,Tzeng2017,Liu2018}.
Compared with the above two groups of methods which are based on latent features, generative-model-based approaches align the source and the target domain by image-to-image transformation which uses a cycle-consistency constraint to preserve domain-invariant features~\cite{Bousmalis2017,Kim2017b,Hoffman2018}.

\begin{figure}[t]
 \centering
  \includegraphics[width=0.49\textwidth, trim=5.5cm 7.8cm 10.5cm 1cm, clip]{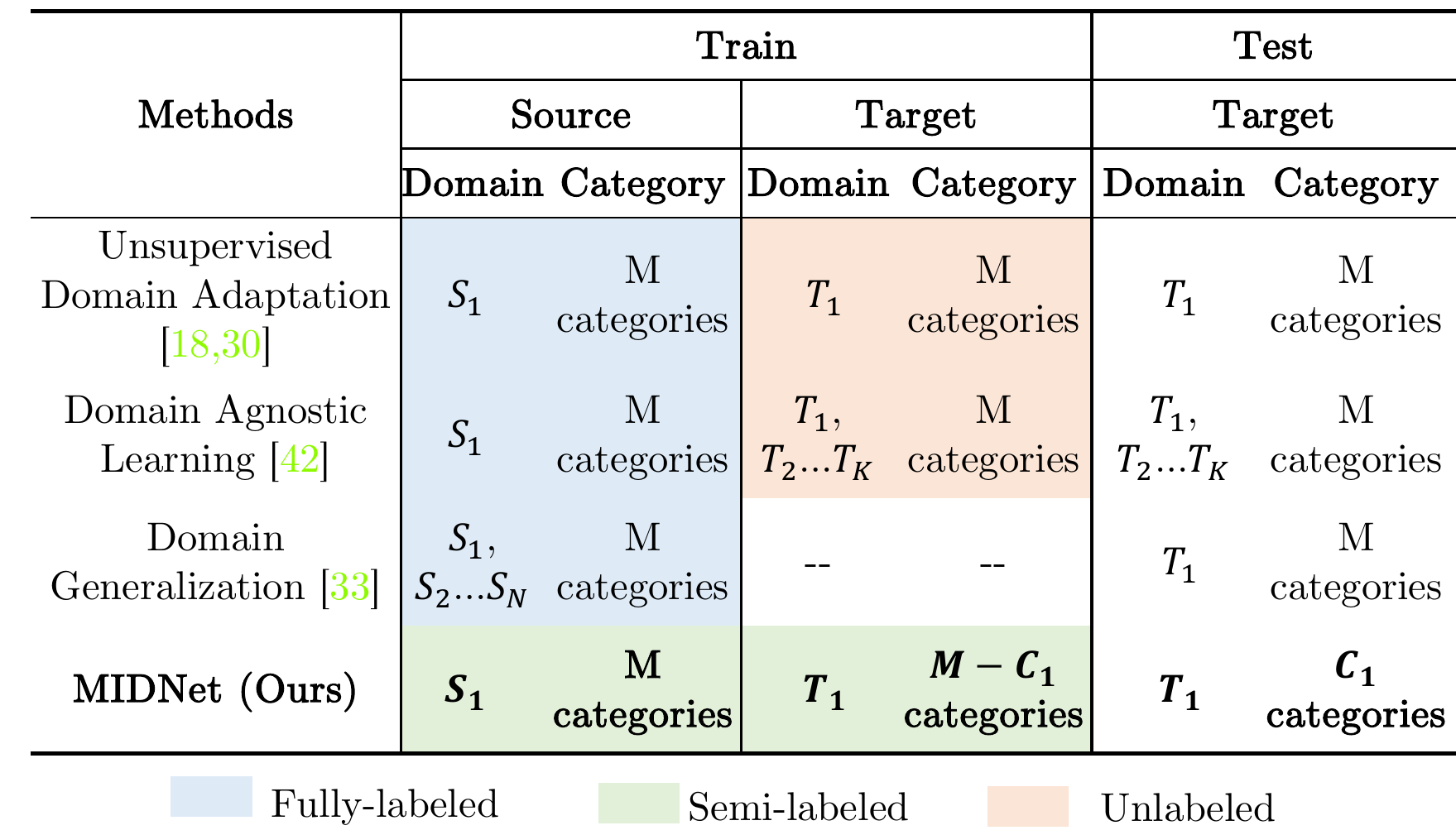}
 \caption{The differences between our method and other existing domain adaptation methods. We mainly compare two aspects, the problem setup and the training paradigm.}
 \label{method_comp}
\end{figure}

However, existing domain adaptation methods can be practically prohibitive in real applications because a large amount of labeled data is needed. Unsupervised domain adaption requires plenty of labeled samples from the source domain~\cite{Lee2019}. 
Domain generalization which has been developed to learn universal knowledge for unseen target domains require many labeled samples from multiple source domains~\cite{LiH2018,LiY2018b}. 
Although adversarial adaptation alternatives can perform well, optimizing adversarial objectives remains difficult and unstable in practice~\cite{Lezama2019}. Most importantly, previous methods make no explicit attempt to disentangle domain-invariant features from domain features, which results in the inability of dealing with previously unseen categories in the target domain. 

We postulate that domain adaptation should be able to learn generalizable features to transfer the knowledge from known entangled image features (e.g. categories from both domains) to new entangled image features (e.g. categories from target domain but not available during training). The difference of task setup between our work and other domain adaptation work is shown in Fig.~\ref{method_comp}.





In this paper, we propose mutual-information-based disentangled networks (MIDNet) for representation disentanglement to address the problem outlined in Fig.~\ref{flowchart}. 
In contrast to previous methods, the proposed approach extracts generalized categorical features by explicitly disentangling entangled image features (categorical features and domain features) via mutual information minimization~\cite{Belghazi2018}. To enhance disentanglement, we estimate the similarity of categorical features from both domains to achieve domain-invariant features, instead of optimizing an adversarial loss. Our method is a semi-supervised learning method, which only requires a small number of labeled samples during training while unlabeled data is integrated using a strategy similar to the MixMatch approach~\cite{Berthelot2019}.

\begin{figure}[tb]
 \centering
 \setcounter{subfigure}{0}
 \subfloat[Digits dataset]{
 \begin{tabular}{@{\hspace{-1\tabcolsep}}c@{\hspace{0.3\tabcolsep}}c@{\hspace{0.3\tabcolsep}}c@{\hspace{0.3\tabcolsep}}c@{\hspace{0.3\tabcolsep}}c@{\hspace{0.3\tabcolsep}}c@{\hspace{0.3\tabcolsep}}c@{\hspace{0.3\tabcolsep}}c@{\hspace{0.3\tabcolsep}}c@{\hspace{0.3\tabcolsep}}c@{\hspace{0.3\tabcolsep}}c@{\hspace{0.3\tabcolsep}}c}
  \raisebox{0.5\height}{\rotatebox[origin=c]{90}{\makecell{~\scalebox{0.7}{MNIST}}}} &
  \includegraphics[height=0.7cm]{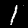} &
  \includegraphics[height=0.7cm]{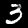} &
  \includegraphics[height=0.7cm]{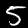} &
  \includegraphics[height=0.7cm]{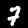} &
  \includegraphics[height=0.7cm]{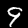} &
  \raisebox{0.4\height}{\rotatebox[origin=c]{90}{\makecell{~\scalebox{0.7}{MNIST-M}}}} &
  \includegraphics[height=0.7cm]{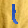} &
  \includegraphics[height=0.7cm]{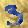} &
  \includegraphics[height=0.7cm]{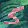} &
  \includegraphics[height=0.7cm]{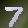} &
  \includegraphics[height=0.7cm]{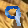}
  \end{tabular}
  }
  \\
  \setcounter{subfigure}{1}
  \subfloat[Fetal US dataset]{
   \begin{tabular}{@{\hspace{-1\tabcolsep}}c@{\hspace{0.3\tabcolsep}}c@{\hspace{0.3\tabcolsep}}c@{\hspace{0.3\tabcolsep}}c@{\hspace{0.3\tabcolsep}}c@{\hspace{0.3\tabcolsep}}c}
  \raisebox{1.5\height}{\rotatebox[origin=c]{90}{\makecell{~\scalebox{0.7}{SF}}}} &
  \includegraphics[height=1.2cm]{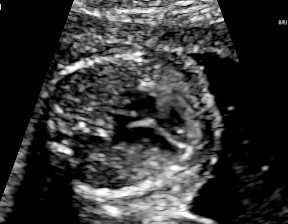} &
  \includegraphics[height=1.2cm]{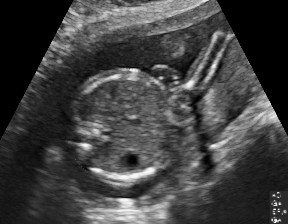} &
  \includegraphics[height=1.2cm]{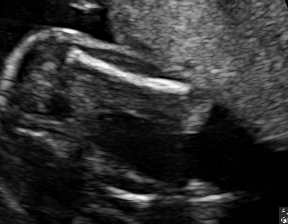} &
  \includegraphics[height=1.2cm]{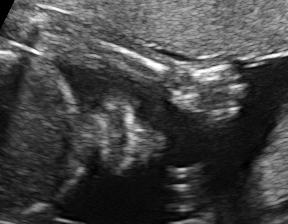} &
  \includegraphics[height=1.2cm]{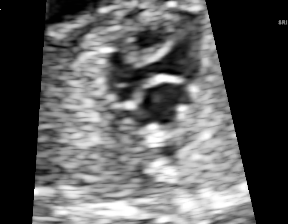} \\
  \raisebox{1.5\height}{\rotatebox[origin=c]{90}{\makecell{~\scalebox{0.7}{SC}}}} &
  \stackunder{\includegraphics[height=1.27cm, trim=2.4cm 1.4cm 2.4cm 1.4cm, clip]{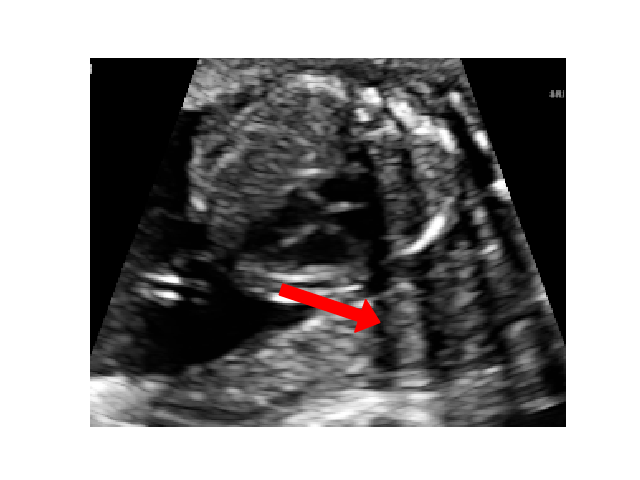}}{~\scalebox{0.7}{4CH}} &
  \stackunder{\includegraphics[height=1.27cm, trim=2.4cm 1.4cm 2.4cm 1.4cm, clip]{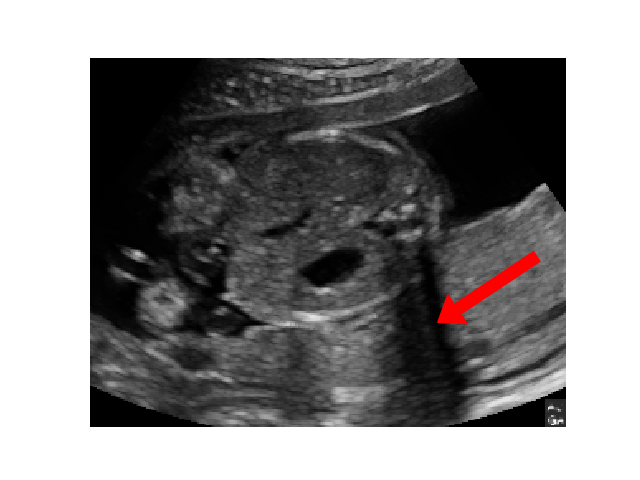}}{~\scalebox{0.7}{Abdominal}} &
  \stackunder{\includegraphics[height=1.27cm, trim=2.4cm 1.4cm 2.4cm 1.4cm, clip]{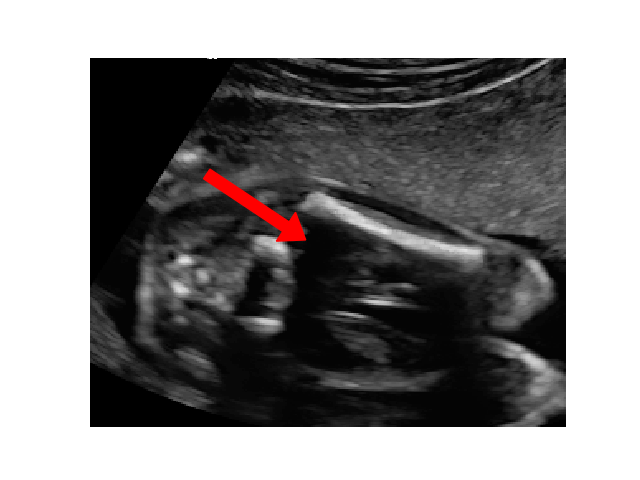}}{~\scalebox{0.7}{Femur}} &
  \stackunder{\includegraphics[height=1.27cm, trim=2.4cm 1.4cm 2.4cm 1.4cm, clip]{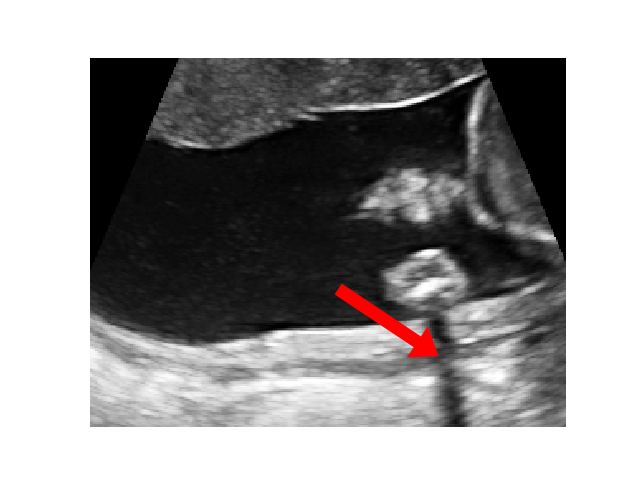}}{~\scalebox{0.7}{Lips}} &
  \stackunder{\includegraphics[height=1.27cm, trim=2.4cm 1.4cm 2.4cm 1.4cm, clip]{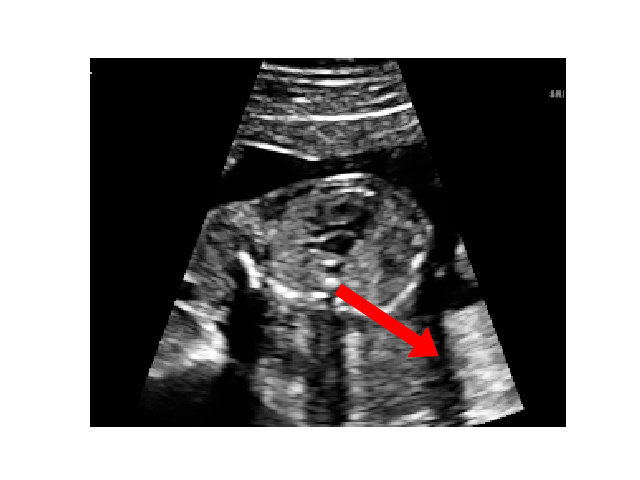}}{~\scalebox{0.7}{LVOT}}
  \end{tabular}
 }
 \\
 \setcounter{subfigure}{2}
  \subfloat[Cross-device dataset]{
   \begin{tabular}{@{\hspace{-0.3\tabcolsep}}c@{\hspace{0.6\tabcolsep}}c@{\hspace{0.6\tabcolsep}}c@{\hspace{0.6\tabcolsep}}c@{\hspace{0.6\tabcolsep}}c@{\hspace{0.6\tabcolsep}}c}
  \raisebox{0.7\height}{\rotatebox[origin=c]{90}{\makecell{~\scalebox{0.8}{Device A}}}} &
  \stackunder{\includegraphics[height=1.3cm]{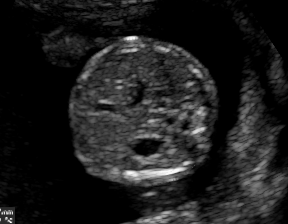}}{~\scalebox{0.8}{Abdominal}} &
  \stackunder{\includegraphics[height=1.3cm]{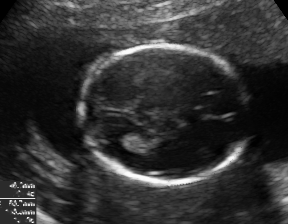}}{~\scalebox{0.8}{Brain}} &
  \raisebox{0.7\height}{\rotatebox[origin=c]{90}{\makecell{~\scalebox{0.8}{Device B}}}} &
  \stackunder{\includegraphics[height=1.3cm]{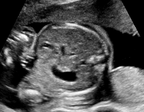}}{~\scalebox{0.8}{Abdominal}} &
  \stackunder{\includegraphics[height=1.3cm]{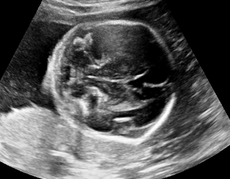}}{~\scalebox{0.8}{Brain}}
  \end{tabular}
 }
  \caption{Examples of datasets. (a) The digits dataset, containing MNIST~\cite{LeCun1998} and MNIST-M~\cite{Ganin2015}. (b) The fetal US dataset, including \textit{shadow-free} (SF) and \textit{shadow-containing} (SC) images of different anatomical structures. Red arrows show examples of acoustic shadow artifacts. (c) The fetal US dataset with images acquired by different devices, device A (GE Voluson E8) and device B ( Philips EPIQ V7 G).
  }
  \label{DataPresentation}
\end{figure}

We demonstrate the efficiency of our method on handwritten digits classification task.
To illustrate practical applicability, we additionally evaluate the proposed method on a challenging medical application, the detection of standardized fetal ultrasound (US) views during prenatal screening. In many countries, this is clinical routine for abnormality screening during pregnancy.
Early detection of pathological development can inform treatment and delivery options~\cite{salomon2011}. DNNs have shown promising performance to support this task~\cite{Baumgartner2017}. 
However, labeled training data is often insufficient as annotating medical images requires significant expertise and is prohibitively expensive in both time and labor. Operator skill dependent manifestation of acoustic shadows~\cite{feldman2005,meng2018} as shown in Fig.~\ref{DataPresentation}(b) as well as different imaging devices as shown in Fig.~\ref{DataPresentation}(c) can lead to a domain shift problem for vanilla DNN classifiers. 
In this work, we explore the domain shift problem respectively caused by shadow artifacts and different imaging devices in the fetal US standard plane classification task.

\noindent
\subsection{Contribution}
The main contributions of this paper are: 
(1) we propose end-to-end trainable Mutual-Information-based Disentangled Networks (MIDNet) for learning generalized features to tackle data with unseen entangled image features, 
(2) we develop MIDNet as a semi-supervised learning paradigm by integrating unlabeled data from both source and target domain to alleviate the demand of annotated data, and (3)we utilize our method for anatomical classification in fetal US, which, to our knowledge, is the first exploration of transferring knowledge to unseen data in a practical application in medical imaging. Our source code will be publicly available at the time of the conference. 

\section{Related work}
\noindent\textbf{Representation disentanglement.} 
Disentangling representations is to interpret underlying interacted factors within data~\cite{Bengio2013,Chen2017} and enable manipulating relevant representations for specific tasks via deep learning to models~\cite{Garcia2018,LiuA2018,Hadad2018}.  Traditional models for this task include techniques such as Independent Component Analysis (ICA)~\cite{Hyvarinen2000} and bilinear models~\cite{Tenenbaum2000} as well as learning-based models such as InfoGAN~\cite{Chen2016} and $\beta$-VAE~\cite{Higgins2017,Burgess2018}.
Recent work by Mathieu et al.~\cite{Mathieu2016} proposes a conditional generative model to disentangle the latent representations into specified and unspecified factors of variation via adversarial training. 
For the same task, Hadad et al.~\cite{Hadad2018} proposes a simpler two-step method with a new adversarial technique for more efficient learning of various unspecified features. Their method directly utilizes the encoded latent space for unspecified factors instead of assuming the underlying distribution.
To use disentangled representations for identifying images with unseen entangled features in real applications, Meng et al.~\cite{meng2019} proposed to disentangle category and domain-specific features using an adversarial regularization in a multi-task learning framework. 
In contrast to these supervised learning methods, we propose an improved model, which evaluates mutual information between latent features to disentangle representations. Additionally, our proposed method uses sparsely labeled data during training.    

\noindent\textbf{Semi-supervised learning (SSL).} The goal of SSL is to address the scarcity of labeled data by leveraging unlabeled data. Various approaches have been proposed for SSL~\cite{Chapelle2006, Lee2013, Laine2016, Miyato2018, Zhang2018, ZhangG2019}. Recently, Zhang et al.~\cite{Zhang2018} proposed a MixUp learning principle to train a model on convex combinations  of samples and their corresponding labels. This principle encourages the model to favor linear behavior between samples and alleviates the problems arising from mislabelled examples. Extending this work, Berthelot et al.~\cite{Berthelot2019} introduced a SSL method, MixMatch, which estimates the low-entropy labels for unlabeled samples and then applies MixUp to mixed labeled and unlabeled samples for training the model. In this paper, we utilize MixMatch to integrate unlabeled samples from both source and target domain during training (Sec.~\ref{mixmat}).

\section{MIDNet: Mutual-Information-based Disentangled Neural Networks}

Our goal is to disentangle categorical features from domain features to obtain generalizable features, so that our model can classify the categories in the target domain which have not been seen during training.
We formulate our task as follows: let $\mathcal{X}^S={\{\mathbf{x}_i^S\}}_{i=1}^{|C^S|}$ be the images from a source domain which contain categories $C^S$ and $\mathcal{X}^T={\{\mathbf{x}_i^T\}}_{i=1}^{|C^T|}$ be images from a target domain with categories $C^T,C^T \subset C^S$.
In both domains, categorical labels are available for part of the images as $\mathcal{Y}^S, \mathcal{Y}^T$. We want to train a network to maximize the categorical prediction performance of the classifier on images in the target domain from new categories ${\{\mathbf{x}^T|\mathbf{x}^T \in C^S-C^T\}}$. 

\begin{figure*}[tb]
 \centering
 \includegraphics[width=\textwidth, trim=8cm 10cm 8cm 3.2cm, clip]{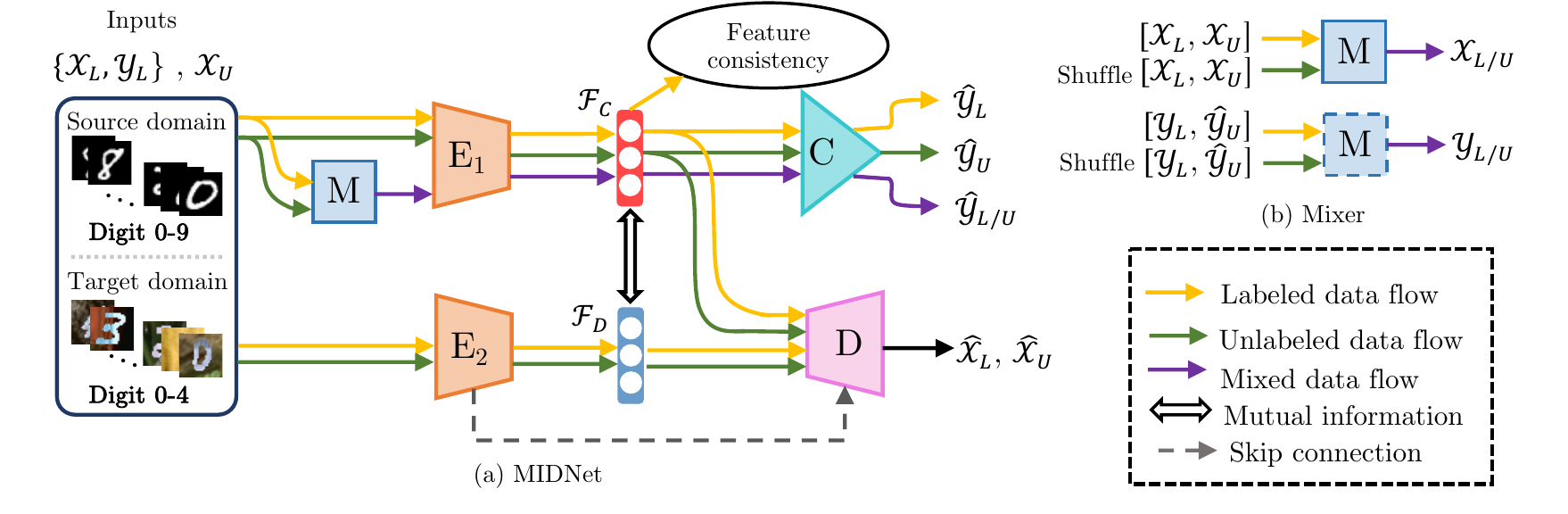}
 \caption{(a) The architecture of the proposed MIDNet. We extract disjoint features ($\mathcal{F}_C, \mathcal{F}_D$) through mutual information minimization in the latent space and apply a feature consistency constraint to extract domain-invariant features $\mathcal{F}_C$ for further disentanglement. $E_1, E_2$ are separate encoders, $C$ is a classifier and $D$ is a decoder. (b) We integrate unlabeled data by a mixer $M$ for semi-supervised learning.
 }
 \label{methed_outline}
\end{figure*}

To solve this task, we propose MIDNet in combination with semi-supervised learning. The architecture of our model is shown in Fig.~\ref{methed_outline}. Two independent encoders $E$ are utilized to respectively extract categorical features $\mathcal{F}_C$ and domain features $\mathcal{F}_D$ from labeled data $\{\mathcal{X}_L, \mathcal{Y}_L\}=\{\mathbf{x}_i|\mathbf{x}_i\in\mathcal{X}^S\cup\mathcal{X}^T, y_i|y_i\in\mathcal{Y}^S\cup\mathcal{Y}^T\}$ and unlabeled data $\mathcal{X}_U=\{\mathbf{x}_j|\mathbf{x}_j\in\mathcal{X}^S\cup\mathcal{X}^T\}$. The class discriminator $C$ is responsible for predicting class distributions from $\mathcal{F}_C$ for both $\mathcal{X}_L$ and $\mathcal{X}_U$ while the decoder $D$ combines $\mathcal{F}_C$ and $\mathcal{F}_D$ for the reconstruction of input images. The mixer $M$ aims to linearly mix labeled and unlabeled samples so that the model is trained to show linear behavior between samples for further leveraging unlabeled data. For representation disentanglement, mutual information between $\mathcal{F}_C$ and $\mathcal{F}_D$ is minimized to encourage $\mathcal{F}_C$ to become domain-invariant and maximally informative for categorical classification. Feature consistency between labeled images is additionally kept to promote the independence of $\mathcal{F}_C$.

\subsection{Image reconstruction}

The first step of MIDNet is to employ an Encoder-Decoder framework for independent extraction of two internal representations from the input data $\mathbf{x}$. Two encoders $E_1, E_2$ are built to respectively generate latent vectors that aim to represent categorical features $\mathcal{F}_C$ and domain features $\mathcal{F}_D$, where $\mathcal{F}_C=E_1(\mathbf{x};\phi_1)$ and $\mathcal{F}_D=E_2(\mathbf{x};\phi_2)$. The decoder $D$ is utilized to guarantee that the combination of these features is capable of recovering original input data, where $\widehat{\mathbf{x}}=D(\mathcal{F}_C, \mathcal{F}_D;\psi)$. Here, $\mathbf{x}\in\mathcal{X}_L\cup\mathcal{X}_U$ and $\phi_1, \phi_2, \psi$ are the parameters of $E_1, E_2, D$, respectively. The cost function of this reconstruction is
\begin{equation}\label{Loss_reconstuction}
\mathcal{L}_{rec}=\| \widehat{x}-x\|_F^2.
\end{equation}

We employ skip connection between $E_2$ and $D$ to integrate high-frequency features from $E_2$ into the reconstruction, which helps $\mathcal{F}_D$ to contain valid information instead of invalid noise.
This image reconstruction extracts two groups of features from internal representations of original data.
The rest of our networks are designed and trained to enable $\mathcal{F}_C$ to only contain categorical information, thus becoming separated from $\mathcal{F}_D$ that only contains domain information.    
 
\subsection{Classification}
\label{cls}

We use a class discriminator $C$ to predict $|C^S|$ labels for labeled data, which encourages $\mathcal{F}_C$ to be maximally informative about categorical classification. $E_1, E_2$ and $C$ are updated by minimizing the cross-entropy loss 
\begin{equation}\label{Loss_classification}
\mathcal{L}_{cls}=-\mathbb{E}_{\{\mathbf{x},y\}\thicksim \{\mathcal{X}_L, \mathcal{Y}_L\}}\sum_{t=1}^{|C^S|}\mathbbm{1}[t=y]log(C(\mathcal{F}_C;\delta)).
\end{equation}
Here $\delta$ refers to the parameters of $C$. At the same time, $C$ predicts the class distribution of the unlabeled data $\mathbf{x} \in \mathcal{X}_U$ as $P_{C}(\widehat{y}|\mathbf{x};\delta), \widehat{y}\in\widehat{\mathcal{Y}}_U$. The predicted class distribution of unlabeled data will be utilized in SSL-based regularization (Sec.~\ref{mixmat}). The class discriminator on its own is unlikely to ensure that categorical features $\mathcal{F}_C$ are domain-invariant. This is because the training objective in Eq.~\ref{Loss_classification} only ensures that $\mathcal{F}_C$ contains as much information as possible for the target classification task.

\subsection{Mutual information disentanglement}

To address the problem from Sect.~\ref{cls}, we minimize the mutual information between $\mathcal{F}_C$ and $\mathcal{F}_D$. This minimization forces $\mathcal{F}_C$ to contain less domain information and thus separates categorical features from domain features.
Mutual information is defined as 
\begin{equation}\label{MI_orig}
I(\mathcal{D}_{\mathcal{F}_C};\mathcal{D}_{\mathcal{F}_D})=\int_{\mathcal{X}\times\mathcal{Z}}log\frac{d\mathbb{P}_{XZ}}{d\mathbb{P}_{X}\bigotimes\mathbb{P}_{Z}}d\mathbb{P}_{XZ}, 
\end{equation}
where $\mathbb{P}_{XZ}$ is the joint probability distribution of $(\mathcal{D}_{\mathcal{F}_C},\mathcal{D}_{\mathcal{F}_D})$, $\mathbb{P}_{X}=\int_\mathcal{Z}d\mathbb{P}_{XZ}$ and $\mathbb{P}_{Z}=\int_\mathcal{X}d\mathbb{P}_{XZ}$ are respectively marginal distributions of $\mathcal{D}_{\mathcal{F}_C}$ and $\mathcal{D}_{\mathcal{F}_D}$.
We utilize Mutual Information Neural Estimation (MINE)~\cite{Belghazi2018} to approximate the lower-bound of mutual information on $n$ samples by a neural network with parameters $\theta\in\Theta$, 
\begin{equation}\label{MINE}
\widehat{I(\mathcal{D}_{\mathcal{F}_C};\mathcal{D}_{\mathcal{F}_D})}_n=\sup_{\theta\in\Theta}E_{\mathbb{P}_{XZ}^{(n)}}\left[T_\theta\right]-log(E_{\mathbb{P}_{X}^{(n)}\bigotimes\widehat{\mathbb{P}}_{Z}^{(n)}}\left[e^{T_\theta}\right]).
\end{equation}
Practically, the expectations in Eq.~\ref{MINE} are estimated by Monte-Carlo integration~\cite{Peng2019} with shuffled samples along the batch axis ($\mathcal{F}'_D$), and thus the cost function of the mutual information disentanglement is 
\begin{equation}\label{MINE_shuffle}
\begin{split}
    \mathcal{L}_{MI}&=-\widehat{I(\mathcal{D}_{\mathcal{F}_C};\mathcal{D}_{\mathcal{F}_D})}_n \\
    &=-(\frac{1}{n}\sum_{i=1}^nT(\mathcal{F}_C,\mathcal{F}_D,\theta)
    -log(\frac{1}{n}\sum_{i=1}^ne^{T(\mathcal{F}_C,\mathcal{F}'_D,\theta)})).
\end{split}
\end{equation}
Here, $(\mathcal{F}_C,\mathcal{F}_D)$ are sampled from joint distributions while $(\mathcal{F}_C,\mathcal{F}'_D)$ are sampled from the product of marginal distributions. 

\subsection{Feature consistency}

Extracting disjoint and complementary features by mutual information disentanglement is not enough for obtaining generalizable features since it fails to align features from source domain to target domain. We hypothesize that categorical features of a certain category are supposed to be consistent between different domains. For better disentanglement, we further enhance $\mathcal{F}_C$ to be domain-invariant by minimizing the distance of categorical features between source domain and target domain 
for samples in $\mathcal{X}_L$ with
\begin{equation}\label{Loss_consistency}
\mathcal{L}_{cons}=\frac{1}{|C^T|}\sum_{i=1}^{|C^T|}(\frac{1}{n_{c_i}}\sum_{j=1}^{n_{c_i}} \|f_{c_{ij}}^S - f_{c_{ij}}^T\|_F^2),
\end{equation}
where $n_{c_i}$ is the number of samples in category $c_i$, $c_i\in C_T$ and $f_{c_{ij}}^S, f_{c_{ij}}^T$ are the categorical features of the $j$th sample from category $c_i$ in the source domain and the target domain, respectively. This feature consistency contributes to align source domain to target domain and thus further helps mutual information disentanglement to extract generalized categorical features. 

\subsection{SSL-based regularization}
\label{mixmat}

Conventional fully supervised learning requires a large number of annotated input images with categorical labels and domain labels. However, in practice, labeled data is not easily available at any granularity. 
Berthelot et al.~\cite{Berthelot2019} propose a SSL-based method, MixMatch, integrating unlabeled data during training to reduce the dependency on labeled data. We adopt a simplified MixMatch using a mixer $M$ to leverage unlabeled data.

As shown in Fig.~\ref{methed_outline} (b), $M$ is utilized to linearly combine two random samples $(\mathbf{x}_1, \mathbf{x}_2)$ and their corresponding labels $(y_1, y_2)$ from labeled data and unlabeled data by 
\begin{equation}\label{mixUp}
\begin{split}
    &\mathbf{x}_{mix}=M(\mathbf{x}_1,\mathbf{x}_2;\beta)=\beta \mathbf{x}_1+(1-\beta)\mathbf{x}_2, \\ &y_{mix}=M(y_1,y_2;\beta)=\beta y_1+(1-\beta)y_2.
\end{split}
\end{equation}
where $\beta=max(\xi, 1-\xi)$, $\xi\thicksim Beta(\alpha,\alpha)$. Here, $\mathbf{x}_1\in\mathcal{X}_{cat}$, $\mathbf{x}_2\in\mathcal{X}'_{cat}$. $\mathcal{X}_{cat}=[\mathcal{X}_L,\mathcal{X}_U]$ is the concatenation of $\mathcal{X}_L$ and $\mathcal{X}_U$. $\mathcal{X}'_{cat}$ is the shuffled $\mathcal{X}_{cat}$ along the batch axis. Similarly, $y_1\in\mathcal{Y}_{cat}$ with $\mathcal{Y}_{cat}=[\mathcal{Y}_L,\widehat{\mathcal{Y}}_U]$, and $y_2\in\mathcal{Y}'_{cat}$. Note that $\widehat{\mathcal{Y}}_U$ is the collection of the predicted labels for unlabeled data according to Sec.~\ref{cls}. We denote that $\mathbf{x}_{mix}\in\mathcal{X}_{L/U}$, $y_{mix}\in\mathcal{Y}_{L/U}$. The goal of this SSL-based regularization is to encourage the linear behavior of the class discriminator, and thus the objective function is 
\begin{equation}\label{SSL}
\mathcal{L}_{SSL}=\|y_{mix}-P_C(\widehat{y}_{mix}|\mathbf{x}_{mix};\delta)\|_F^2, 
\end{equation}
where $P_C(\widehat{y}_{mix}|\mathbf{x}_{mix};\delta)=C(\mathbf{x}_{mix};\delta)$ is the predicted label of $\mathbf{x}_{mix}$ via class discriminator $C$.

\subsection{Optimization}

Our model is an end-to-end trainable framework and the overall objective is a linear combination of all cost functions 
\begin{equation}\label{Loss}
min\{\lambda_1\mathcal{L}_{rec}+\lambda_2\mathcal{L}_{cls}+\lambda_3\mathcal{L}_{MI}+\lambda_4\mathcal{L}_{cons}+\lambda_5\mathcal{L}_{SSL}\}, 
\end{equation}
where $\lambda_1$ to $\lambda_5$ are hyper-parameters chosen experimentally depending on the dataset. We optimize the MINE and the rest of our model in an alternative fashion. Inspired by~\cite{Belghazi2018}, we use the Adam optimizer ($\text{beta}=0.9$, $\text{learning rate}=10^{-5}$) to train the network parameters $\theta$ based on Eq.~\ref{MINE_shuffle} and use Stochastic Gradient Descent (SGD) with momentum optimizer ($\text{momentum}=0.9$, $\text{learning rate}=10^{-5}$) to update the parameters of encoders, decoders and class discriminator based on Eq.~\ref{Loss}.
We apply L2 regularization ($\text{scale}=10^{-5}$) to all weights during training to prevent over-fitting and we apply random image flipping as data augmentation. 
Classes are kept balanced on labeled data during training. Our model is trained on a Nvidia Titan X GPU with 12 GB of memory.

\begin{table*}[htb]
\centering
\begin{threeparttable}
\caption{Comparison of baselines and ablation study (MIDNet-I to MIDNet-VIII) for \textbf{digit classification task}. $30\%$ of training data are labeled data and the rest are unlabeled data. Average F1-score, Recall and Precision are measured on three groups of test data. Best results are shown in bold.}
\label{mnist_table}
\begin{tabular}{cccc|ccc|ccc}
\toprule
\toprule
\multirow{2}{*}{Methods}                            & 
\multicolumn{3}{c|}{$T_{Source}$}                    &
\multicolumn{3}{c|}{$T_{Target}$}                    &
\multicolumn{3}{c}{$T_{Target}^{New}$}              \\
\cmidrule{2-10}
~~~~~                                               &
F1-score                                                & 
Recall                                              & 
Precision                                           &
F1-score                                                & 
Recall                                              & 
Precision                                           &
F1-score                                                & 
Recall                                              & 
Precision                                           \\
\midrule
Source only                                   &
0.9253                                              &
0.9254                                              &
0.9256                                              &
0.5309                                              &
0.5293                                              &
0.5340                                              &
0.5114                                              &
0.5118                                              &
0.5213                                              \\
VGG~\cite{Simonyan15}                                                 &
0.9151                                              &
0.9162                                              &
0.9146                                              &
0.7334                                              &
0.8412                                              &
0.6517                                              &
0.6152                                              &
0.5208                                              &
0.7552                                              \\
Res-VGG~\cite{Simonyan15,He2016}                    &
0.9802                                              &
0.9802                                              &
0.9802                                              &
0.7236                                              &
0.9338                                              &
0.5953                                              &
0.5228                                              &
0.3595                                              &
0.9631                                              \\
Two-step-fair~\cite{Hadad2018}                                       &
0.8704                                              &
0.8707                                              &
0.8704                                             &
0.6908                                              &
0.7806                                              &
0.6203                                              &
0.5794                                              &
0.5002                                              &
0.6911                                              \\
Two-step-Unfair~\cite{Hadad2018}                                       &
0.7465                                              &
0.7492                                              &
0.7591                                             &
0.5839                                              &
0.6407                                              &
0.5428                                              &
0.2983                                              &
0.2598                                              &
0.3894                                              \\
Multi-task~\cite{meng2019}                                          &
0.9318                                              &
0.9315                                              &
0.9332                                             &
0.6203                                              &
0.7824                                              &
0.5171                                              &
0.5053                                              &
0.3695                                              &
0.8368                                              \\
DANN~\cite{Ganin2016}                                          &
0.9678                                              &
0.9679                                              &
0.9681                                              &
0.6818                                              &
0.8901                                              &
0.5579                                              &
0.4506                                              &
0.3023                                              &
0.9091                                              \\
MME~\cite{Saito2019}                                          &
0.9709                                              &
0.9704                                              &
0.9726
&
0.7357                                              &
0.9426                                              &
0.6205                                              &
0.4858                                              &
0.3287                                              &
0.9722                                              \\
\midrule
MIDNet-I                                           &
0.9836                                              &
0.9837                                              &
0.9835                                             &
0.7039                                              &
0.9115                                              &
0.5797                                              &
0.4956                                              &
0.3376                                              &
0.9431                                              \\
MIDNet-II                                           &
0.9841                                              &
0.9842                                              &
0.9842                                             &
0.7059                                              &
0.9160                                              &
0.5809                                              &
0.4916                                              &
0.3322                                              &
0.9501                                              \\
MIDNet-III                                           &
0.9869                                              &
0.9869                                              &
0.9870                                             &
\textbf{0.8333}                                              &
0.9780                                              &
\textbf{0.7298}                                              &
\textbf{0.7511}                                              &
\textbf{0.6137}                                              &
0.9765                                              \\
MIDNet-IV                                           &
0.9858                                              &
0.9860                                              &
0.9859                                             &
0.7439                                              &
0.9569                                              &
0.6169                                              &
0.5207                                              &
0.3566                                              &
0.9771                                              \\
MIDNet-V                                           &
0.9863                                              &
0.9862                                              &
0.9864                                             &
0.8051                                              &
0.9766                                              &
0.6903                                              &
0.6821                                              &
0.5295                                              &
\textbf{0.9807}                                              \\
MIDNet-VI                                           &
0.9868                                              &
0.9869                                              &
0.9868                                             &
0.7532                                              &
0.9602                                              &
0.6253                                              &
0.5541                                              &
0.3900                                              &
0.9689                                              \\
MIDNet-VII                                           &
0.9881                                              &
0.9881                                              &
0.9881                                             &
0.8223                                              &
0.9779                                              &
0.7140                                              &
0.7280                                              &
0.5820                                              &
0.9791                                              \\
MIDNet-VIII                                          &
\textbf{0.9906}                                              &
\textbf{0.9905}                                              &
\textbf{0.9906}                                             &
0.8204                                              &
\textbf{0.9803}                                              &
0.7108                                              &
0.7166                                              &
0.5704                                              &
0.9806                                              \\
\bottomrule
\bottomrule
\end{tabular}
\begin{tablenotes}
\item The baselines and ablation study models are introduced in Sec.~\ref{experiment}.
\end{tablenotes}
\end{threeparttable}
\end{table*}

\section{Experiments}
\label{experiment}
We evaluate the proposed method on three tasks, digit classification with digit recognition benchmarks (MNIST~\cite{LeCun1998} and MNIST-M~\cite{Ganin2015}, Fig.~\ref{DataPresentation}(a)) and two standard plane classification tasks with medical fetal US images (Fig.~\ref{DataPresentation}(b),(c)). 

We compare the proposed method with the state-of-the-art algorithms that can be used for the main task of this work. We explore the effectiveness of different components in MIDNet via an ablation study. By training MIDNet with different percentage of labeled data, we also evaluate the performance of our model in semi-supervised setting.

We utilize three groups of test data for the evaluation: (1) test data from the source domain $T_{Source}$, e.g. digits 0 to 9 in MNIST (Fig.~\ref{methed_outline} (a)), (2) test data from the target domain whose image attributes have been observed during training $T_{Target}$ , e.g. digits 0 to 4 in MNIST-M (Fig.~\ref{methed_outline} (a)), and (3), most importantly, test data from the target domain whose image attributes are absent during training $T_{Target}^{New}$ , e.g. digits 5 to 9 in MNIST-M. 
We show the major results in the main paper and detailed implementation as well as more results in the supplementary material.

\noindent\textbf{Comparison methods.} We evaluate a VGG network~\cite{Simonyan15} which is trained on data only from the source domain, namely \textit{Source only}, as a baseline to demonstrate that the domain shift problems affects the generalizability of deep models. To verify that MIDNet is able to extract generalized features across domains, we compare MIDNet with a VGG network~\cite{Simonyan15} and
a VGG network with residual unit~\cite{He2016} (Res-VGG). We further compare MIDNet to the state-of-the-art feature disentanglement algorithms for addressing the task in this work, including a two-step disentanglement method~\cite{Hadad2018} and a multi-task learning based disentanglement method~\cite{meng2019}. Note that we implement the method in~\cite{Hadad2018} differently from the original paper. Specifically, we train the model simultaneously to enable it to be suitable for our task setup. We denote \textit{Two-step-fair} as~\cite{Hadad2018} with an adversarial network using unspecific features ($Z$) for category classification and denote \textit{Two-step-Unfair} as~\cite{Hadad2018} with an adversarial network using specific features ($S$) for domain classification.  
We keep the original experimental settings for the method in~\cite{meng2019} (namely Multi-task). All comparison methods above are fully-supervised. Additionally, we compare the proposed method with the state-of-the-art domain adaptation methods, including domain-adversarial training of neural networks (DANN)~\cite{Ganin2016} and semi-supervised domain adaptation via minimax Eentropy (MME)~\cite{Saito2019}. These two comparison methods are semi-supervised. 

\noindent\textbf{Ablation study.} For the ablation study, we remove different loss components to obtain different combinations of components in MIDNet. MIDNet-I: only contains classification and reconstruction; MIDNet-II: MIDNet-I plus mutual information disentanglement; MIDNet-III: MIDNet-I plus feature consistency; MIDNet-IV: MIDNet-I plus SSL based regularization; MIDNet-V: MIDNet-II plus feature consistency; MIDNet-VI: MIDNet-III plus SSL-based regularization;  MIDNet-VII: MIDNet-IV plus feature consistency; MIDNet-VIII: contains all components.

\begin{figure}[tb]
 \centering
 \begin{tabular}{c@{\hspace{0.3\tabcolsep}}c@{\hspace{0.3\tabcolsep}}c}
 ~~~~~  &
 \raisebox{0.2\height}{\rotatebox[origin=c]{0}{\makecell{~\scalebox{1}{\textbf{Semi-supervised}}}}}  &
 \raisebox{0.2\height}{\rotatebox[origin=c]{0}{\makecell{~\scalebox{1}{\textbf{Fully-supervised}}}}}  \\
 \raisebox{2.6\height}{\rotatebox[origin=c]{90}{\makecell{~\scalebox{1}{$\mathbf{T_{Source}}$}}}} &
  \includegraphics[height=4cm, trim=2cm 0cm 3cm 0cm, clip]{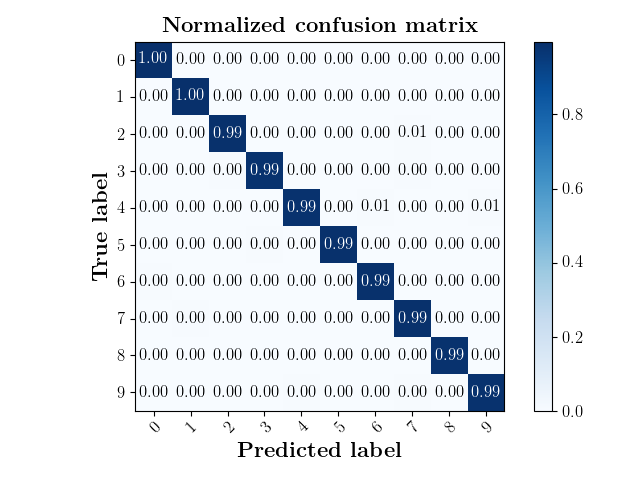} &
  \includegraphics[height=4cm, trim=2cm 0cm 1.4cm 0cm, clip]{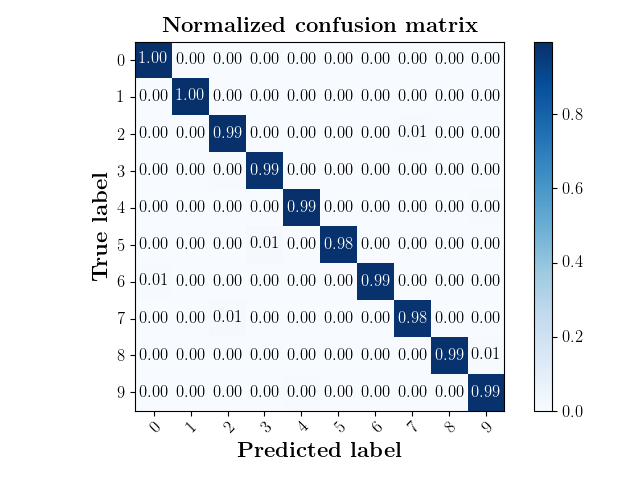} \\
  \raisebox{1.2\height}{\rotatebox[origin=c]{90}{\makecell{~\scalebox{1}{$\mathbf{T_{Target}}$ $\&$ $\mathbf{T_{Target}^{New}}$}}}} &
  \includegraphics[height=4cm, trim=2cm 0cm 3cm 0cm, clip]{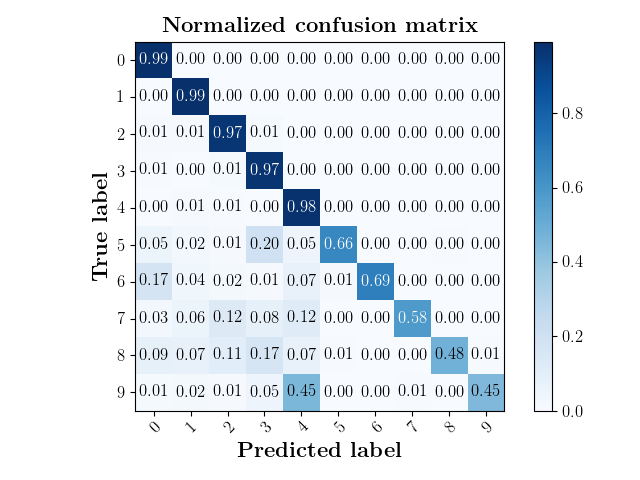} &
  \includegraphics[height=4cm, trim=2cm 0cm 1.4cm 0cm, clip]{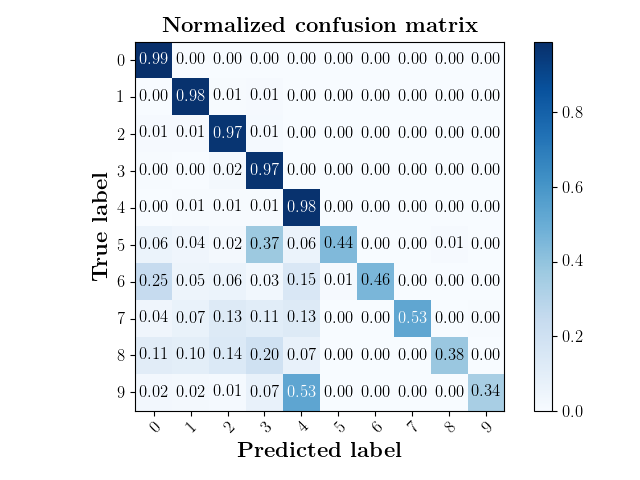}
  \end{tabular}
  \caption{Confusion matrices for digit classification: semi-supervised setting versus fully-supervised setting with MIDNet-VIII model as backbone. For the semi-supervised setting, $30\%$ of training data are labeled data and the rest are unlabeled data. The fully-supervised learning in this experiment only uses the $30\%$ labeled data for training, without using unlabeled data.}
  \label{Mnist_CM}
\end{figure}

\subsection{Experiments on digit data}
In this experiment, MNIST is the source domain while MNIST-M is the target domain. Except \textit{Source only}, all the methods are trained on digits 0 to 9 from the source domain and digits 0 to 4 from the target domain. We aim to separate digital features (categorical features) from domain features to obtain generalized digital features, and thus to achieve high digit classification performance on $T_{Target}^{New}$ (digits 5 to 9 from target domain). Here, $T_{Source}$ contains digits 0 to 9 from the source domain and $T_{Target}$ contains digits 0 to 4 from the target domain.
Hyper-parameters $\lambda_1$ to $\lambda_5$ in Eq.~\ref{Loss} are experimentally chosen as $\lambda_1=1, \lambda_2=10, \lambda_3=10^{-3}, \lambda_4=10^{2}, \lambda_5=10^{3}$. 

\noindent\textbf{Results.}
The experimental results of baselines and the ablation study are shown in Table.~\ref{mnist_table}. From this table, we observe that the MIDNet-VIII model significantly outperforms other baselines on all test data for average F1-score, recall and precision. For example, MIDNet-VIII achieves average F1-score of $0.9906$, $0.8204$ and $0.7166$ for $T_{Source}$, $T_{Target}$ and $T_{Target}^{New}$, respectively, while the highest average F1-score of other baselines on the corresponding test data are $0.9802$ (Res-VGG~\cite{Simonyan15,He2016}), $0.7357$ (MME~\cite{Saito2019}) and $0.5794$ (Two-step-fair~\cite{Hadad2018}). Additionally, MIDNet-III performs slightly better than MIDNet-VIII on $T_{Target}$ and $T_{Target}^{New}$, demonstrating that feature consistency is important for digit classification. The results of MIDNet-IV and MIDNet-I (similarly, MIDNet-VI vs. MIDNet-II and MIDNet-VIII vs. MIDNet-V) illustrate the effectiveness of SSL based regularization in the proposed MIDNet.

We further compare the performance of MIDNet-VIII in a semi-supervised setting and a fully-supervised setting. Here, the semi-supervised setting utilizes the training data containing $30\%$ labeled data and $70\%$ unlabeled data, while the fully supervised setting only uses the $30\%$ labeled data. The confusion matrix in Fig.~\ref{Mnist_CM} shows the effectiveness of unlabeled data in our proposed method, for example, the classification accuracy of $T_{Target}^{New}$ greatly improves when integrating unlabeled data (semi-supervised).

\begin{figure}[tb]
 \centering
 \subfloat[Digit classification]{
  \includegraphics[height=3.5cm, trim=0cm 0cm 1.2cm 0.2cm, clip]{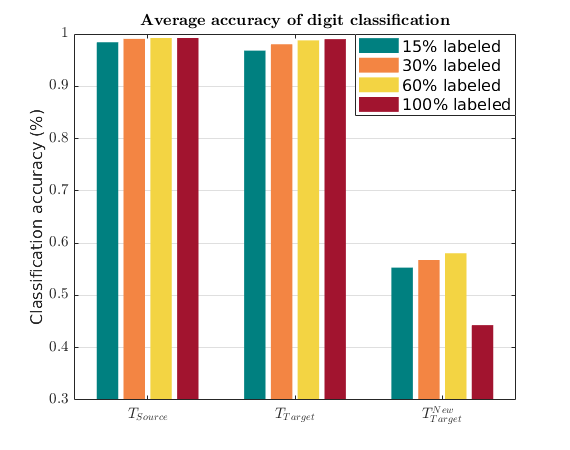}
  }
  \hfill
  \hspace{-3ex}
  \subfloat[Standard plane classification]{
  \includegraphics[height=3.5cm, trim=0cm 0cm 1.2cm 0.2cm, clip]{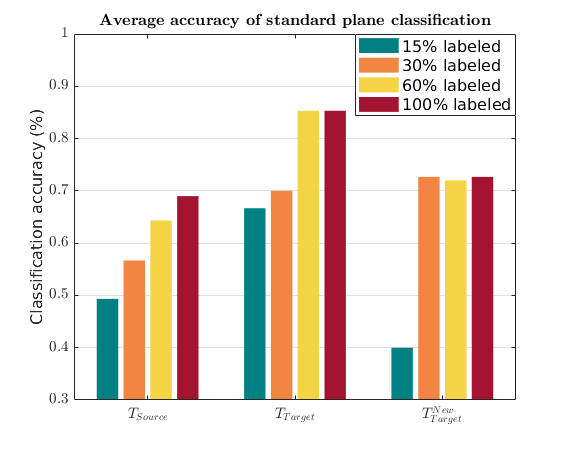}
  }
  \caption{Average accuracy of (a) digit classification and (b) fetal US standard plane classification with different percentage of labeled data ($15\%, 30\%, 60\%, 100\%$) for semi-supervised learning based on MIDNet-VIII model.}
  \label{semi}
\end{figure}

To explore the importance of labeled data, we evaluate the performance of MIDNet-VIII based on using $15\%, 30\%, 60\%$ and $100\%$ labeled data during training. Fig.~\ref{semi} (a) shows the average accuracy of these experiments on three groups of test data. From this figure, we observe that the classification performance only slightly improves with increasing labeled data. This indicates that MIDNet is capable of achieving expected performance with sparsely labeled data. 

\subsection{Experiments on ultrasound data}
The fetal US dataset consists of $\sim 7k$ 2D fetal US images sampled from 2694 2D US examinations with gestational ages between $18-22$ weeks. Eight different US systems of identical make and model (GE Voluson E8) were used for the acquisitions to eliminate as many unknown image acquisition parameters as possible. Six different anatomical standard plane locations have been selected by expert sonographers, including Four Chamber View (4CH), Abdominal, Femur, Lips, Left Ventricular Outflow Tract (LVOT) and Right Ventricular Outflow Tract (RVOT). 
The images have additionally been classified by expert observers as shadow-containing or shadow-free.
In this experiment, the source domain contains shadow-free images (see Fig.~\ref{DataPresentation} (b) SF) while the target domain has shadow-containing images from less skilled sonographers and less favorable imaging conditions (see Fig.~\ref{DataPresentation} (b) SC). Training data consists of all six standard planes from the source domain as well as Abdominal, LVOT and RVOT from the target domain. We aim to separate anatomical features (categorical features) and shadow artifacts features (domain features) to obtain generalized anatomical features for achieving high performance of standard plane classification on $T_{Target}^{New}$ (4CH, Femur and Lips from target domain). Here, $T_{Source}$ contains 4CH, Abdominal, Femur, Lips, LVOT and RVOT from the source domain and $T_{Target}$ contains Abdominal, LVOT and RVOT from the target domain.
Hyper-parameters $\lambda_1$ to $\lambda_5$ in Eq.~\ref{Loss} are $\lambda_1=1, \lambda_2=10, \lambda_3=10^{-4}, \lambda_4=10, \lambda_5=10$.

\begin{table*}[htb]
\centering
\caption{Comparison of baselines and ablation study (MIDNet-I to MIDNet-VIII) for the \textbf{fetal US standard plane classification task}. $30\%$ of training data are labeled data and the rest are unlabeled data. Average F1-score, Recall and Precision are measured on three groups of test data. Best results are shown in bold.}
\label{US_table}
\begin{tabular}{cccc|ccc|ccc}
\toprule
\toprule
\multirow{2}{*}{Methods}                            & 
\multicolumn{3}{c|}{$T_{Source}$}                    &
\multicolumn{3}{c|}{$T_{Target}$}                    &
\multicolumn{3}{c}{$T_{Target}^{New}$}              \\
\cmidrule{2-10}
~~~~~                                               &
F1-score                                                & 
Recall                                              & 
Precision                                           &
F1-score                                                & 
Recall                                              & 
Precision                                           &
F1-score                                                & 
Recall                                              & 
Precision                                           \\
\midrule
Source only                                   &
\textbf{0.5558}                                              &
0.5600                                              &
0.5590                                              &
0.4882                                              &
0.4600                                              &
0.5285                                              &
0.5675                                              &
0.5867                                              &
0.5629                                              \\
VGG~\cite{Simonyan15}                                                 &
0.5440                                              &
0.5467                                              &
0.5433                                              &
0.4096                                              &
0.4267                                              &
0.3987                                              &
0.5261                                              &
0.5200                                              &
0.5326                                              \\
Res-VGG~\cite{Simonyan15,He2016}                    &
0.4354                                              &
0.4500                                              &
0.4793                                              &
0.5377                                              &
0.5800                                              &
0.5148                                              &
0.6100                                              &
0.5867                                              &
0.6607                                              \\
Two-step-fair~\cite{Hadad2018}                                       &
0.4531                                              &
0.4500                                              &
0.4572                                             &
0.4400                                              &
0.4467                                              &
0.4338                                              &
0.5008                                              &
0.4933                                              &
0.5095                                              \\
Two-step-Unfair~\cite{Hadad2018}                                       &
0.4894                                              &
0.4933                                              &
0.4895                                             &
0.4515                                              &
0.4733                                              &
0.4319                                              &
0.4571                                              &
0.4400                                              &
0.4769                                              \\
Multi-task~\cite{meng2019}                                          &
0.4622                                              &
0.4667                                              &
0.5524                                             &
0.5787                                              &
0.5667                                              &
0.6220                                              &
0.6393                                              &
0.6533                                              &
0.6491                                              \\
\midrule
MIDNet-I                                           &
0.4643                                              &
0.4767                                              &
0.5891                                             &
0.5796                                              &
0.5933                                              &
0.5944                                              &
0.6280                                              &
0.6133                                              &
0.6947                                              \\
MIDNet-II                                           &
0.4760                                              &
0.4867                                              &
0.5336                                             &
0.6185                                              &
0.6533                                              &
0.6056                                              &
0.6559                                              &
0.6200                                              &
0.7412                                              \\
MIDNet-III                                           &
0.4929                                              &
0.5100                                              &
0.5498                                             &
0.5620                                              &
0.5800                                              &
0.5512                                              &
0.6887                                              &
0.6667                                              &
0.7267                                              \\
MIDNet-IV                                           &
0.4636                                              &
0.4833                                              &
0.5403                                             &
0.5746                                              &
0.5867                                              &
0.5705                                              &
0.6378                                              &
0.6400                                              &
0.6732                                              \\
MIDNet-V                                           &
0.5379                                              &
0.5533                                              &
0.6007                                             &
0.5976                                              &
0.6600                                              &
0.5612                                              &
0.6603                                              &
0.6000                                              &
0.8119                                              \\
MIDNet-VI                                           &
0.4195                                              &
0.4367                                              &
0.5102                                             &
0.5657                                              &
0.5800                                              &
0.5637                                              &
0.6025                                              &
0.6000                                              &
0.6539                                              \\
MIDNet-VII                                           &
0.5339                                              &
0.5467                                              &
0.5948                                             &
0.6654                                              &
\textbf{0.7067}                                              &
0.6449                                              &
0.7091                                              &
0.6600                                              &
\textbf{0.8255}                                              \\
MIDNet-VIII                                           &
0.5484                                              &
\textbf{0.5667}                                              &
\textbf{0.6683}                                             &
\textbf{0.6809}                                              &
0.7000                                              &
\textbf{0.6803}                                              &
\textbf{0.7399}                                              &
\textbf{0.7267}                                              &
0.7830                                              \\
\bottomrule
\bottomrule
\end{tabular}
\end{table*}

\noindent\textbf{Results.}
Table.~\ref{US_table} shows experimental results of the baselines as well as the ablation study of the proposed MIDNet for the fetal US standard plane classification. From this table, we observe that the MIDNet-VIII model significantly outperforms all the baselines on $T_{Target}$ and ${T_{Target}^{New}}$.
Interestingly, \textit{Source only} achieves best average F1-score ($+0.0074$ better than MIDNet-VIII) on $T_{Source}$. This is somewhat expected as \textit{Source only} is not disturbed by the target domain.
Among all the models in the ablation study, MIDNet-VIII achieves the best performance in most cases. The results of MIDNet-VII and MIDNet-VIII demonstrate that mutual information disentanglement is essential.


From Fig.~\ref{semi} (b), we observe that classification performance improves with the increase of labeled data in most cases. However, the performance reaches its peak after a certain percentage of labeled data is added. For example, the saturation point is $60\%$ for $T_{Target}$ and $30\%$ for $T_{Target}^{New}$ in this experiment. This indicates that semi-supervised learning 
is beneficial for achieving expected performance with sparsely labeled data. 

\noindent\textbf{Cross-device experiment.} The previous experiment on fetal US images is supported by data restricted to one US imaging device. We here evaluate MIDNet for a standard plane classification task on data from different imaging devices (different device domains).
The source domain contains Abdominal, Brain, Femur and Lips from device A (GE Voluson E8) while the target domain includes Abdominal and Brain from device B (Philips EPIQ V7 G) during training. $T_{Source}$ consists of Abdominal, Brain, Femur and Lips from device A, $T_{Target}$ contains Abdominal and Brain from device B and $T_{Target}^{New}$ has Femur and Lips from device B. Hyper-parameters $\lambda_1$ to $\lambda_5$ in Eq.~\ref{Loss} are $\lambda_1=1, \lambda_2=10, \lambda_3=10^{-4}, \lambda_4=50, \lambda_5=50$. The average classification performance of baselines and the MIDNet-VIII model are shown in Table.~\ref{iFind12_table}. We observe that our method significantly outperforms the state-of-the-art on $T_{Target}$ and most importantly on $T_{Target}^{New}$. DANN~\cite{Ganin2016} is the best on $T_{Source}$ because of the provided target labels during training.


\begin{table*}[tb]
\centering
\caption{Comparison of \textit{Source only}, domain adaptation methods and MIDNet-VIII for the \textbf{fetal US standard plane classification task with data from different acquisition devices} (Cross-device experiment). $30\%$ of training data are labeled data and the rest are unlabeled data. Best results in bold.}
\label{iFind12_table}
\resizebox{\textwidth}{!}{
\begin{tabular}{cccc|ccc|ccc}
\toprule[1.2pt]
\multirow{2}{*}{Methods}                            & 
\multicolumn{3}{c|}{$T_{Source}$}                    &
\multicolumn{3}{c|}{$T_{Target}$}                    &
\multicolumn{3}{c}{$T_{Target}^{New}$}              \\
\cmidrule{2-10}
~~~~~                                               &
F1-score                                                & 
Recall                                              & 
Precision                                           &
F1-score                                                & 
Recall                                              & 
Precision                                           &
F1-score                                                & 
Recall                                              & 
Precision                                           \\
\midrule
Source only                                   &
0.7665                                              &
0.7700                                              &
0.7264                                              &
0.6971                                              &
0.6750                                              &
0.7305                                              &
0.6742                                              &
0.7050                                              &
0.6899                                              \\
DANN~\cite{Ganin2016}                                          &
\textbf{0.9699}                                              &
\textbf{0.9700}                                             &
\textbf{0.9704}                                             &
0.3568                                              &
0.4800                                              &
0.3253                                              &
0.3470                                              &
0.3400                                             &
0.3542                                              \\
MME~\cite{Saito2019}                                          &
0.9526                                              &
0.9525                                              &
0.9537
&
0.5400
&
0.7150                                               &
0.4345                                              &
0.4293                                              &
0.3600 
&
\textbf{0.9595}                                              \\
MIDNet-VIII (Ours)                                  &
0.9281                                              &
0.9275                                             &
0.9327                                             &
\textbf{0.7434}                                              &
\textbf{0.7300}                                             &
\textbf{0.7676}                                             &
\textbf{0.8383}                                              &
\textbf{0.8600}                                             &
0.8497                                             \\
\bottomrule[1.2pt]
\end{tabular}
}
\end{table*}


\section{Discussion}
\label{discuss}
The performance of semi-supervised learning usually positively correlates with the percentage of labeled data. In our experimental setting, excessive labeled data may lead to increased class imbalance. This issue may result in decreased classification performance as shown in the quantitative results of $T_{Target}^{New}$ in Fig.~\ref{semi} (a).

Natural and medical images contain complex entangled image features. For example, shadow artifacts in US imaging are caused by anatomies through blocking the propagation of sound waves or destructive interference. 
Traditional DNN-based classifiers jointly learn shadow features and anatomical features without understanding the underlying semantics. 
By observing classification performance on source and target domain after separating entangled image attributes, our model can be potentially used to interpret the effective factors for target tasks. 
For example, the results of the classification performance of $T_{Source}$ and $T_{Target}$ (Table.~\ref{US_table}) indicate that shadow features can be more informative for some categories than the actual anatomy.

In this work, we only compare our method with unsupervised/semi-supervised domain adaptation methods that aim at one source/target domain. This is because other types of domain adaptation methods (shown in Fig.~\ref{flowchart}(b)) have different task setup from this work (e.g. domain agnostic learning requires multiple source domains while domain generalization requires multiple target domains).

One limitation of the proposed method is that it requires hyper-parameters to be fine-tuned depending on the overarching application. Ideas from meta-learning~\cite{feurer2015efficient} will be explored in future work to allow more efficient deployment. 
A limitation specific to this paper is that the proposed method is evaluated on only one target domain. When more domains are available, our model can be extended to address unseen categories from multiple target domains by adding more encoder-decoder pairs. Furthermore, we only apply our method to image classification tasks. Other applications such as image segmentation and object detection could be explored by utilizing a decoder instead of a classifier in the model.

\section{Conclusion}
In this paper, we discuss a problem that is rarely evaluated but important in practical scenarios: transferring knowledge from known entangled image features (e.g. categorical features and domain features) to unseen entangled image features (e.g. categories
from target domain but are not available during training). We propose Mutual-Information-based Disentangled Neural Networks (MIDNet) to extract generalizable features. Our model is developed with a semi-supervised learning paradigm. Experiments on handwritten digits data and fetal US images demonstrate the efficiency and practical applicability of our method compared with the state-of-the-art.

{\small
\bibliographystyle{unsrtnat}

\begin{thebibliography}{56}
\providecommand{\natexlab}[1]{#1}
\providecommand{\url}[1]{\texttt{#1}}
\expandafter\ifx\csname urlstyle\endcsname\relax
  \providecommand{\doi}[1]{doi: #1}\else
  \providecommand{\doi}{doi: \begingroup \urlstyle{rm}\Url}\fi

\bibitem[Quinonero-Candela et~al.(2008)Quinonero-Candela, Sugiyama,
  Schwaighofer, and Lawrence]{joaquin2009}
Joaquin Quinonero-Candela, Masashi Sugiyama, Anton Schwaighofer, and Neil~D.
  Lawrence.
\newblock \emph{Dataset Shift in Machine Learning}.
\newblock Neural Information Processing. MIT Press, 2008.
\newblock ISBN 9780262170055.

\bibitem[Saenko et~al.(2010)Saenko, Kulis, Fritz, and Darrell]{Saenko2010}
Kate Saenko, Brian Kulis, Mario Fritz, and Trevor Darrell.
\newblock Adapting visual category models to new domains.
\newblock In \emph{Proceedings of the 11th European Conference on Computer
  Vision: Part IV}, ECCV'10, pages 213--226, Berlin, Heidelberg, 2010.
  Springer-Verlag.
\newblock ISBN 3-642-15560-X, 978-3-642-15560-4.

\bibitem[Long et~al.(2015)Long, Cao, Wang, and Jordan]{Long2015}
Mingsheng Long, Yue Cao, Jianmin Wang, and Michael~I. Jordan.
\newblock Learning transferable features with deep adaptation networks.
\newblock In \emph{Proceedings of the 32Nd International Conference on
  International Conference on Machine Learning - Volume 37}, ICML'15, pages
  97--105. JMLR.org, 2015.

\bibitem[Tzeng et~al.(2017)Tzeng, Hoffman, Saenko, and Darrell]{Tzeng2017}
Eric Tzeng, Judy Hoffman, Kate Saenko, and Trevor Darrell.
\newblock Adversarial discriminative domain adaptation.
\newblock pages 2962--2971, 07 2017.

\bibitem[Zhang et~al.(2017{\natexlab{a}})Zhang, David, and Gong]{Zhang2017}
Yang Zhang, Philip David, and Boqing Gong.
\newblock Curriculum domain adaptation for semantic segmentation of urban
  scenes.
\newblock \emph{ICCV'2017}, pages 2039--2049, 2017{\natexlab{a}}.

\bibitem[Zou et~al.(2018)Zou, Yu, Kumar, and Wang]{Zou2018}
Yang Zou, Zhiding Yu, B.~V. K.~Vijaya Kumar, and Jinsong Wang.
\newblock Unsupervised domain adaptation for semantic segmentation via
  class-balanced self-training.
\newblock In \emph{ECCV}, 2018.

\bibitem[Dou et~al.(2019)Dou, de~Castro, Kamnitsas, and Glocker]{Dou2019}
Qi~Dou, Daniel~Coelho de~Castro, Konstantinos Kamnitsas, and Ben Glocker.
\newblock Domain generalization via model-agnostic learning of semantic
  features.
\newblock In \emph{NeurIPS'2019}, 2019.

\bibitem[Chartsias et~al.(2019)Chartsias, Joyce, Papanastasiou, Semple,
  Williams, Newby, Dharmakumar, and Tsaftaris]{Chartsias2019}
Agisilaos Chartsias, Thomas Joyce, Giorgos Papanastasiou, Scott Semple,
  Michelle Williams, David Newby, Rohan Dharmakumar, and Sotirios Tsaftaris.
\newblock Disentangled representation learning in cardiac image analysis.
\newblock \emph{Medical Image Analysis}, 58:\penalty0 101535, 07 2019.

\bibitem[Saito et~al.(2018)Saito, Watanabe, Ushiku, and Harada]{Saito2018}
Kuniaki Saito, Kohei Watanabe, Yoshitaka Ushiku, and Tatsuya Harada.
\newblock Maximum classifier discrepancy for unsupervised domain adaptation.
\newblock In \emph{The IEEE Conference on Computer Vision and Pattern
  Recognition (CVPR)}, June 2018.

\bibitem[Lee et~al.(2019)Lee, Batra, Baig, and Ulbricht]{Lee2019}
Chen-Yu Lee, Tanmay Batra, Mohammad~Haris Baig, and Daniel Ulbricht.
\newblock Sliced wasserstein discrepancy for unsupervised domain adaptation.
\newblock In \emph{Proceedings of the IEEE Conference on Computer Vision and
  Pattern Recognition}, pages 10285--10295, 2019.

\bibitem[Kamnitsas et~al.(2017)Kamnitsas, Baumgartner, Ledig, Newcombe,
  Simpson, Kane, Menon, Nori, Criminisi, Rueckert, et~al.]{Kamnitsas2016}
Konstantinos Kamnitsas, Christian Baumgartner, Christian Ledig, Virginia
  Newcombe, Joanna Simpson, Andrew Kane, David Menon, Aditya Nori, Antonio
  Criminisi, Daniel Rueckert, et~al.
\newblock Unsupervised domain adaptation in brain lesion segmentation with
  adversarial networks.
\newblock In \emph{International conference on information processing in
  medical imaging}, pages 597--609. Springer, 2017.

\bibitem[Chen et~al.(2019{\natexlab{a}})Chen, Dou, Chen, Qin, and
  Heng]{Chen2019}
Cheng Chen, Qi~Dou, Hao Chen, Jing Qin, and Pheng{-}Ann Heng.
\newblock Synergistic image and feature adaptation: Towards cross-modality
  domain adaptation for medical image segmentation.
\newblock In \emph{The Thirty-Third {AAAI} Conference on Artificial
  Intelligence, {AAAI}, Honolulu, Hawaii, USA, January 27 - February 1, 2019.},
  pages 865--872, 2019{\natexlab{a}}.

\bibitem[Geirhos et~al.(2018)Geirhos, Rubisch, Michaelis, Bethge, Wichmann, and
  Brendel]{geirhos2018imagenet}
Robert Geirhos, Patricia Rubisch, Claudio Michaelis, Matthias Bethge, Felix~A.
  Wichmann, and Wieland Brendel.
\newblock Imagenet-trained cnns are biased towards texture; increasing shape
  bias improves accuracy and robustness.
\newblock \emph{arXiv:1811.12231}, 2018.

\bibitem[Peng et~al.(2019)Peng, Huang, Sun, and Saenko]{Peng2019}
Xingchao Peng, Zijun Huang, Ximeng Sun, and Kate Saenko.
\newblock Domain agnostic learning with disentangled representations.
\newblock In \emph{ICML}, 2019.

\bibitem[Tzeng et~al.(2014)Tzeng, Hoffman, Zhang, Saenko, and
  Darrell]{Tzeng2014}
Eric Tzeng, Judy Hoffman, Ning Zhang, Kate Saenko, and Trevor Darrell.
\newblock Deep domain confusion: Maximizing for domain invariance.
\newblock \emph{arXiv:1412.3474}, 12 2014.

\bibitem[Sun et~al.(2016)Sun, Feng, and Saenko]{Sun2016}
Baochen Sun, Jiashi Feng, and Kate Saenko.
\newblock Return of frustratingly easy domain adaptation.
\newblock In \emph{Proceedings of the Thirtieth AAAI Conference on Artificial
  Intelligence}, AAAI'16, pages 2058--2065. AAAI Press, 2016.

\bibitem[Cao et~al.(2018)Cao, Long, Wang, and Jordan]{Cao2019}
Zhangjie Cao, Mingsheng Long, Jianmin Wang, and Michael~I. Jordan.
\newblock Partial transfer learning with selective adversarial networks.
\newblock In \emph{CVPR'18}, pages 2724--2732. {IEEE} Computer Society, 2018.

\bibitem[Chen et~al.(2019{\natexlab{b}})Chen, Wang, Long, and
  Wang]{ChenXinyang2019}
Xinyang Chen, Sinan Wang, Mingsheng Long, and Jianmin Wang.
\newblock Transferability vs. discriminability: Batch spectral penalization for
  adversarial domain adaptation.
\newblock In Kamalika Chaudhuri and Ruslan Salakhutdinov, editors,
  \emph{ICML'2019}, volume~97 of \emph{Proceedings of Machine Learning
  Research}, pages 1081--1090. {PMLR}, 2019{\natexlab{b}}.

\bibitem[Ganin et~al.(2016)Ganin, Ustinova, Ajakan, Germain, Larochelle,
  Laviolette, Marchand, and Lempitsky]{Ganin2016}
Yaroslav Ganin, Evgeniya Ustinova, Hana Ajakan, Pascal Germain, Hugo
  Larochelle, Fran\c{c}ois Laviolette, Mario Marchand, and Victor Lempitsky.
\newblock Domain-adversarial training of neural networks.
\newblock \emph{J. Mach. Learn. Res.}, 17\penalty0 (1):\penalty0 2096--2030,
  January 2016.
\newblock ISSN 1532-4435.

\bibitem[Liu et~al.(2018{\natexlab{a}})Liu, Yeh, Fu, Wang, Chiu, and
  Wang]{Liu2018}
Yen-Cheng Liu, Yu-Ying Yeh, Tzu-Chien Fu, Sheng-De Wang, Wei-Chen Chiu, and
  Yu-Chiang~Frank Wang.
\newblock Detach and adapt: Learning cross-domain disentangled deep
  representation.
\newblock pages 8867--8876, 06 2018{\natexlab{a}}.

\bibitem[Bousmalis et~al.(2016)Bousmalis, Silberman, Dohan, Erhan, and
  Krishnan]{Bousmalis2017}
Konstantinos Bousmalis, Nathan Silberman, David Dohan, Dumitru Erhan, and Dilip
  Krishnan.
\newblock Unsupervised pixel-level domain adaptation with generative
  adversarial networks.
\newblock \emph{2017 IEEE CVPR}, pages 95--104, 2016.

\bibitem[Kim et~al.(2017)Kim, Cha, Kim, Lee, and Kim]{Kim2017b}
Taeksoo Kim, Moonsu Cha, Hyunsoo Kim, Jung~Kwon Lee, and Jiwon Kim.
\newblock Learning to discover cross-domain relations with generative
  adversarial networks.
\newblock In \emph{Proc. 34th ICML'17}, volume~70 of \emph{PMLR}, pages
  1857--1865, Sydney, Australia, 06--11 Aug 2017. PMLR.

\bibitem[Hoffman et~al.(2018)Hoffman, Tzeng, Park, Zhu, Isola, Saenko, Efros,
  and Darrell]{Hoffman2018}
Judy Hoffman, Eric Tzeng, Taesung Park, Jun-Yan Zhu, Phillip Isola, Kate
  Saenko, Alexei Efros, and Trevor Darrell.
\newblock {C}y{CADA}: Cycle-consistent adversarial domain adaptation.
\newblock In \emph{Proc. 35th ICML'18}, volume~80 of \emph{PMLR}, pages
  1989--1998. PMLR, 10--15 Jul 2018.

\bibitem[Li et~al.(2018{\natexlab{a}})Li, Pan, Wang, and Kot]{LiH2018}
Haoliang Li, Sinno~Jialin Pan, Shiqi Wang, and Alex~ChiChung Kot.
\newblock Domain generalization with adversarial feature learning.
\newblock \emph{2018 IEEE/CVF Conference on Computer Vision and Pattern
  Recognition}, pages 5400--5409, 2018{\natexlab{a}}.

\bibitem[Li et~al.(2018{\natexlab{b}})Li, Gong, Tian, Liu, and Tao]{LiY2018b}
Ya~Li, Mingming Gong, Xinmei Tian, Tongliang Liu, and Dacheng Tao.
\newblock Domain generalization via conditional invariant representations.
\newblock In \emph{Thirty-Second AAAI Conference on Artificial Intelligence},
  2018{\natexlab{b}}.

\bibitem[Lezama(2019)]{Lezama2019}
José Lezama.
\newblock Overcoming the disentanglement vs reconstruction trade-off via
  jacobian supervision.
\newblock In \emph{International Conference on Learning Representations}, 2019.

\bibitem[Belghazi et~al.(2018)Belghazi, Baratin, Rajeshwar, Ozair, Bengio,
  Courville, and Hjelm]{Belghazi2018}
Mohamed~Ishmael Belghazi, Aristide Baratin, Sai Rajeshwar, Sherjil Ozair,
  Yoshua Bengio, Aaron Courville, and Devon Hjelm.
\newblock Mutual information neural estimation.
\newblock In \emph{Proc. 35th ICML'18}, volume~80, pages 531--540. PMLR, 10--15
  Jul 2018.

\bibitem[Berthelot et~al.(2019)Berthelot, Carlini, Goodfellow, Papernot,
  Oliver, and Raffel]{Berthelot2019}
David Berthelot, Nicholas Carlini, Ian Goodfellow, Nicolas Papernot, Avital
  Oliver, and Colin Raffel.
\newblock Mixmatch: A holistic approach to semi-supervised learning.
\newblock In \emph{Proc. 33rd NeurIPS}, 05 2019.

\bibitem[LeCun et~al.(1998)LeCun, Bottou, Bengio, Haffner, et~al.]{LeCun1998}
Yann LeCun, L{\'e}on Bottou, Yoshua Bengio, Patrick Haffner, et~al.
\newblock Gradient-based learning applied to document recognition.
\newblock \emph{Proc. IEEE}, 86\penalty0 (11):\penalty0 2278--2324, 1998.

\bibitem[Ganin and Lempitsky(2015)]{Ganin2015}
Yaroslav Ganin and Victor Lempitsky.
\newblock Unsupervised domain adaptation by backpropagation.
\newblock In \emph{Proceedings of the 32Nd International Conference on
  International Conference on Machine Learning - Volume 37}, ICML'15, pages
  1180--1189, 2015.

\bibitem[Salomon et~al.(2011)Salomon, Alfirevic, Berghella, Bilardo,
  Hernandez-Andrade, Johnsen, Kalache, Leung, Malinger, Munoz,
  et~al.]{salomon2011}
L.~J. Salomon, Z.~Alfirevic, V.~Berghella, C.~Bilardo, E.~Hernandez-Andrade,
  S.~L. Johnsen, K.~Kalache, K.‐Y. Leung, G.~Malinger, H.~Munoz, et~al.
\newblock Practice guidelines for performance of the routine mid‐trimester
  fetal ultrasound scan.
\newblock \emph{Ultrasound Obst Gyn}, 37:\penalty0 116--126, 2011.

\bibitem[Baumgartner et~al.(2017)Baumgartner, Kamnitsas, Matthew, Fletcher,
  Smith, Koch, Kainz, and Rueckert]{Baumgartner2017}
C.~Baumgartner, K.~Kamnitsas, J.~Matthew, T.~Fletcher, S.~Smith, L.~Koch,
  B.~Kainz, and D.~Rueckert.
\newblock {SonoNet}: real-time detection and localisation of fetal standard
  scan planes in freehand ultrasound.
\newblock \emph{IEEE Trans. Med. Imaging}, 36\penalty0 (11):\penalty0
  2204--2215, 2017.

\bibitem[Feldman et~al.(2009)Feldman, Katyal, and Blackwood]{feldman2005}
Myra~K. Feldman, Sanjeev Katyal, and Margaret~S. Blackwood.
\newblock Us artifacts.
\newblock \emph{Radio Graphics}, 29:\penalty0 1179–1189, 2009.

\bibitem[Meng et~al.(2019{\natexlab{a}})Meng, Sinclair, Zimmer, Hou, Rajchl,
  Toussaint, Oktay, Schlemper, Gomez, Housden, Matthew, Rueckert, Schnabel, and
  Kainz]{meng2018}
Qingjie Meng, Matthew Sinclair, Veronika Zimmer, Benjamin Hou, Martin Rajchl,
  Nicolas Toussaint, Ozan Oktay, Jo~Schlemper, Alberto Gomez, James Housden,
  Jacqueline Matthew, Daniel Rueckert, Julia~A Schnabel, and Bernhard Kainz.
\newblock Weakly supervised estimation of shadow confidence maps in fetal
  ultrasound imaging.
\newblock \emph{IEEE transactions on medical imaging}, 2019{\natexlab{a}}.
\newblock ISSN 0278-0062.

\bibitem[Bengio et~al.(2013)Bengio, Courville, and Vincent]{Bengio2013}
Yoshua Bengio, Aaron Courville, and Pascal Vincent.
\newblock Representation learning: A review and new perspectives.
\newblock \emph{IEEE Trans. Pattern Anal. Mach. Intell.}, 35\penalty0
  (8):\penalty0 1798--1828, August 2013.
\newblock ISSN 0162-8828.

\bibitem[Chen et~al.(2016{\natexlab{a}})Chen, Kingma, Salimans, Duan, Dhariwal,
  Schulman, Sutskever, and Abbeel]{Chen2017}
Xi~Chen, Diederik~P Kingma, Tim Salimans, Yan Duan, Prafulla Dhariwal, John
  Schulman, Ilya Sutskever, and Pieter Abbeel.
\newblock Variational lossy autoencoder.
\newblock \emph{arXiv preprint arXiv:1611.02731, {ICLR'17}},
  2016{\natexlab{a}}.

\bibitem[Gonzalez-Garcia et~al.(2018)Gonzalez-Garcia, van~de Weijer, and
  Bengio]{Garcia2018}
Abel Gonzalez-Garcia, Joost van~de Weijer, and Yoshua Bengio.
\newblock Image-to-image translation for cross-domain disentanglement.
\newblock In \emph{NeurIPS'18}, pages 1287--1298. Curran Associates, Inc.,
  2018.

\bibitem[Liu et~al.(2018{\natexlab{b}})Liu, Liu, Yeh, and Wang]{LiuA2018}
Alexander~H. Liu, Yen-Cheng Liu, Yu-Ying Yeh, and Yu-Chiang~Frank Wang.
\newblock A unified feature disentangler for multi-domain image translation and
  manipulation.
\newblock In \emph{NeurIPS'18}, pages 2590--2599. Curran Associates, Inc.,
  2018{\natexlab{b}}.

\bibitem[Hadad et~al.(2018)Hadad, Wolf, and Shahar]{Hadad2018}
Naama Hadad, Lior Wolf, and Moni Shahar.
\newblock A two-step disentanglement method.
\newblock In \emph{Proc. IEEE CVPR'18}, pages 772--780, 2018.

\bibitem[Hyv\"{a}rinen and Oja(2000)]{Hyvarinen2000}
A.~Hyv\"{a}rinen and E.~Oja.
\newblock Independent component analysis: Algorithms and applications.
\newblock \emph{Neural Netw.}, 13\penalty0 (4-5):\penalty0 411--430, May 2000.
\newblock ISSN 0893-6080.

\bibitem[Tenenbaum and Freeman(2000)]{Tenenbaum2000}
Joshua~B. Tenenbaum and William~T. Freeman.
\newblock Separating style and content with bilinear models.
\newblock \emph{Neural Comput.}, 12\penalty0 (6):\penalty0 1247--1283, June
  2000.
\newblock ISSN 0899-7667.

\bibitem[Chen et~al.(2016{\natexlab{b}})Chen, Duan, Houthooft, Schulman,
  Sutskever, and Abbeel]{Chen2016}
Xi~Chen, Yan Duan, Rein Houthooft, John Schulman, Ilya Sutskever, and Pieter
  Abbeel.
\newblock Infogan: Interpretable representation learning by information
  maximizing generative adversarial nets.
\newblock In \emph{NeurIPS'16}, pages 2180--2188, USA, 2016{\natexlab{b}}.
  Curran Associates Inc.
\newblock ISBN 978-1-5108-3881-9.

\bibitem[Higgins et~al.(2017)Higgins, Matthey, Pal, Burgess, Glorot, Botvinick,
  Mohamed, and Lerchner]{Higgins2017}
Irina Higgins, Loic Matthey, Arka Pal, Christopher Burgess, Xavier Glorot,
  Matthew Botvinick, Shakir Mohamed, and Alexander Lerchner.
\newblock beta-vae: Learning basic visual concepts with a constrained
  variational framework.
\newblock \emph{ICLR}, 2\penalty0 (5):\penalty0 6, 2017.

\bibitem[Burgess et~al.(2018)Burgess, Higgins, Pal, Matthey, Watters,
  Desjardins, and Lerchner]{Burgess2018}
Christopher~P. Burgess, Irina Higgins, Arka Pal, Lo{\"{\i}}c Matthey, Nick
  Watters, Guillaume Desjardins, and Alexander Lerchner.
\newblock Understanding disentangling in {\(\beta\)}-vae.
\newblock \emph{arXiv:1804.03599}, 2018.

\bibitem[Mathieu et~al.(2016)Mathieu, Zhao, Zhao, Ramesh, Sprechmann, and
  LeCun]{Mathieu2016}
Michael~F Mathieu, Junbo~Jake Zhao, Junbo Zhao, Aditya Ramesh, Pablo
  Sprechmann, and Yann LeCun.
\newblock Disentangling factors of variation in deep representations using
  adversarial training.
\newblock In \emph{NeurIPS'16}, pages 5040--5048, 2016.

\bibitem[Meng et~al.(2019{\natexlab{b}})Meng, Pawlowski, Rueckert, and
  Kainz]{meng2019}
Qingjie Meng, Nick Pawlowski, Daniel Rueckert, and Bernhard Kainz.
\newblock Representation disentanglement for multi-task learning with
  application to fetal ultrasound.
\newblock In \emph{Smart Ultrasound Imaging and Perinatal, Preterm and
  Paediatric Image Analysis}, pages 47--55. Springer, 2019{\natexlab{b}}.

\bibitem[Chapelle et~al.(2006)Chapelle, Schölkopf, and Zien]{Chapelle2006}
Olivier Chapelle, Bernhard Schölkopf, and Alexander Zien.
\newblock \emph{Semi-Supervised Learning}.
\newblock MIT Press, 2006.
\newblock ISBN 9780262033589.

\bibitem[Lee(2013)]{Lee2013}
Dong-Hyun Lee.
\newblock Pseudo-label : The simple and efficient semi-supervised learning
  method for deep neural networks.
\newblock \emph{ICML 2013 Workshop : Challenges in Representation Learning
  (WREPL)}, 07 2013.

\bibitem[Laine and Aila(2017)]{Laine2016}
Samuli Laine and Timo Aila.
\newblock Temporal ensembling for semi-supervised learning.
\newblock In \emph{ICLR'17}, 2017.

\bibitem[Miyato et~al.(2018)Miyato, ichi Maeda, Koyama, and Ishii]{Miyato2018}
Takeru Miyato, Shin ichi Maeda, Masanori Koyama, and Shin Ishii.
\newblock Virtual adversarial training: A regularization method for supervised
  and semi-supervised learning.
\newblock \emph{IEEE Transactions on Pattern Analysis and Machine
  Intelligence}, 41:\penalty0 1979--1993, 2018.

\bibitem[Zhang et~al.(2017{\natexlab{b}})Zhang, Cisse, Dauphin, and
  Lopez-Paz]{Zhang2018}
Hongyi Zhang, Moustapha Cisse, Yann Dauphin, and David Lopez-Paz.
\newblock mixup: Beyond empirical risk minimization.
\newblock In \emph{ICLR'18}, 2017{\natexlab{b}}.

\bibitem[Zhang et~al.(2019)Zhang, Wang, Xu, and Grosse]{ZhangG2019}
Guodong Zhang, Chaoqi Wang, Bowen Xu, and Roger Grosse.
\newblock Three mechanisms of weight decay regularization.
\newblock In \emph{International Conference on Learning Representations}, 2019.

\bibitem[Simonyan and Zisserman(2015)]{Simonyan15}
Karen Simonyan and Andrew Zisserman.
\newblock Very deep convolutional networks for large-scale image recognition.
\newblock In \emph{International Conference on Learning Representations}, 2015.

\bibitem[He et~al.(2016)He, Zhang, Ren, and Sun]{He2016}
Kaiming He, Xiangyu Zhang, Shaoqing Ren, and Jian Sun.
\newblock Deep residual learning for image recognition.
\newblock \emph{2016 IEEE CVPR'16}, pages 770--778, 2016.

\bibitem[Saito et~al.(2019)Saito, Kim, Sclaroff, Darrell, and
  Saenko]{Saito2019}
Kuniaki Saito, Donghyun Kim, Stan Sclaroff, Trevor Darrell, and Kate Saenko.
\newblock Semi-supervised domain adaptation via minimax entropy.
\newblock In \emph{ICCV'19}, 2019.

\bibitem[Feurer et~al.(2015)Feurer, Klein, Eggensperger, Springenberg, Blum,
  and Hutter]{feurer2015efficient}
Matthias Feurer, Aaron Klein, Katharina Eggensperger, Jost Springenberg, Manuel
  Blum, and Frank Hutter.
\newblock Efficient and robust automated machine learning.
\newblock In \emph{Advances in neural information processing systems}, pages
  2962--2970, 2015.

\end{thebibliography}

}

\newpage
\appendix
\noindent\textbf{\huge{Appendices}}

\section{Comparison with current literature}
\begin{figure*}[htb]
 \centering
 \includegraphics[width=\textwidth, trim=3cm 6cm 7cm 3cm, clip]{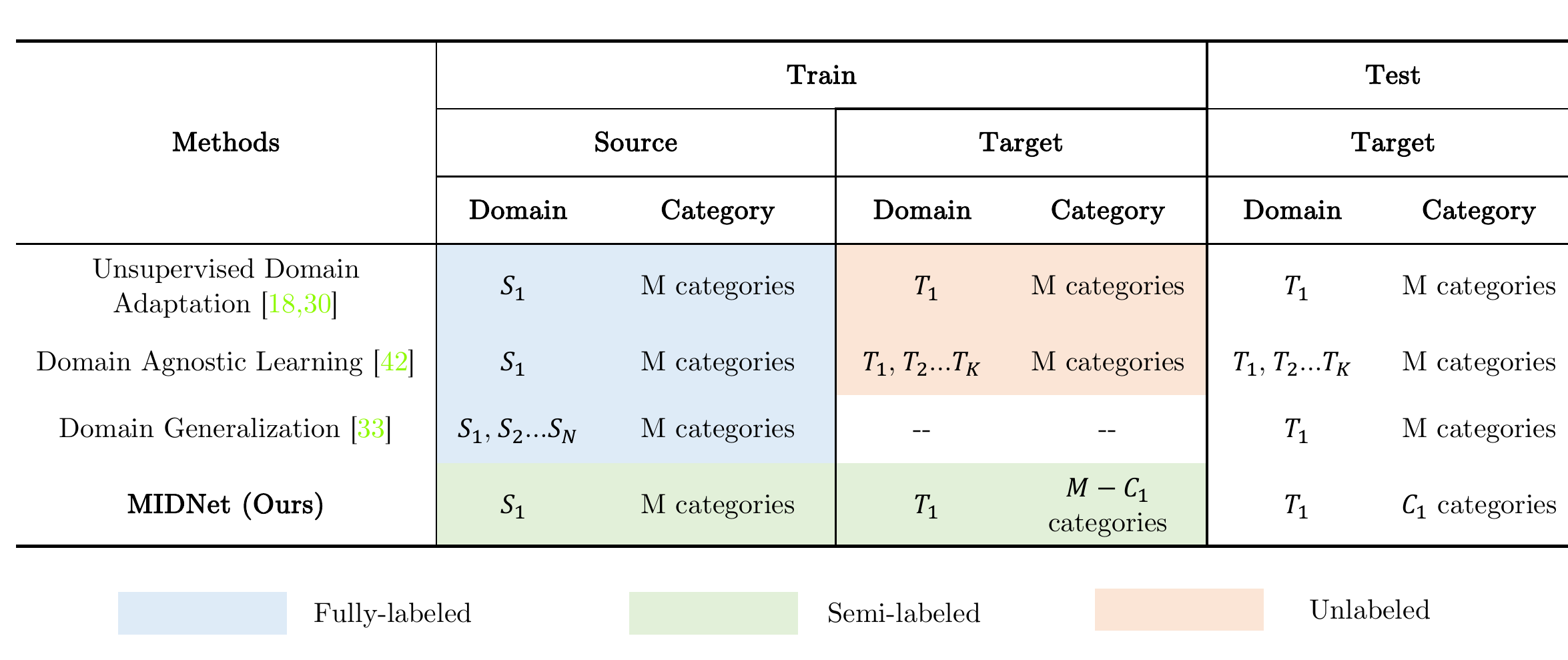}
 \caption{The differences between the proposed method (MIDNet) and other existing domain adaptation methods as they are textually described in the main paper. We mainly compare two aspects, the target problem and the training paradigm.}
 \label{methodComp}
\end{figure*}

In this section, we expand on the main paper's analysis about the differences between our method and various existing groups of domain adaptation methods on two aspects: target problem and training paradigm. Fig.~\ref{methodComp} shows that, (1) from target problem prospective, our method addresses the problem of classification on unseen categories in the target domain while other literature aims at categories in the target domain which have been seen during training, (2) for the training paradigm, our method is semi-supervised in both source and target domains while other literature is fully-supervised in the source domain and unsupervised in the target domain.

\section{Network architectures}

We use Python-inspired pseudo code to present the detailed network architecture of MIDNet. The architecture shows the parameter settings (e.g. filters, strides, hidden units) in the cross-device experiment. 
The implementation of the {\fontfamily{qcr}\selectfont residual\_def} has been integrated from the publicly available DLTK framework~\footnote{https://dltk.github.io/}. Our implementation is on Tensorflow. The parameter settings for the other experiments are shown in Table.~\ref{paraSet}.

\begin{table}[htb]
\centering
\caption{The parameter settings of the MIDNet architecture for different experiments. $N$ is the batch size. Digits refers to the handwritten digits classification task that separates digit features from domain features. Fetal US refers to the experiment that disentangles anatomies from shadow artifacts for fetal US standard plane classification.}
\label{paraSet}
\begin{tabular}{ccc}
\toprule
\toprule
~~~~~                            & 
Digits                    & 
Fetal US                   \\
\midrule
Input dimension                   &
($N$, 28, 28, 3, 1)             &
($N$, 224, 288, 1, 1)             \\
Filters ($E_1/E_2$)                                   &
(8, 16, 32, 8)                                             &
(8, 16, 32, 64, 8)                                             \\
Strides ($E_1/E_2$)                             &
(1, 2, 2, 1)                                     &
(1, 2, 2, 2, 1)                                     \\
Units ($C$)                                  &
(128, 128)                                              &
(128, 128)                                             \\
\bottomrule
\bottomrule
\end{tabular}
\end{table}

\section{Details of data split}

In this section, we provide detailed train/validation/test data split for three experiments on two different datasets. Table.~\ref{DigitSplit} shows the MNIST/MINIST-M split for digits classification. Table.~\ref{USSplitSA} is the fetal US data split for standard plane classification that \textbf{separate anatomical features from artifacts features}. Table.~\ref{USSplitCrossdivece} is the fetal US data split for standard plane classification that \textbf{disjoin anatomical features from imaging devices features}. 

\begin{table*}[htb]
\centering
\caption{Data split of digits classification task. \textcolor{cyan}{Blue} test data is $T_{Source}$, \textcolor{green}{Green} test data is $T_{Target}$ and \textcolor{red}{red} test data is $T_{Target}^{New}$. In this case, all labeled data is $\sim30\%$ of all training data. Train:validation is about 8:2.}
\label{DigitSplit}
\begin{tabular}{cccccccccccc}
\toprule
\toprule
~~~~~             &
\multicolumn{11}{c}{MNIST (Source domain)}                           \\
\midrule[1.5pt]
~~~~~                    & 
0                   &
1                   &
2                  &
3                   &
4                   &
5                   &
6                   &
7                   &
8                   &
9                   &
Unlabeled           \\
Train                    & 
1087                   &
1087                 &
1087                  &
1087                   &
1087                  &
2174                   &
2174                   &
2174                   &
2174                   &
2174                   &
31690           \\
Validation                    & 
1185                   &
1349                    &
1192                  &
1227                   &
1169                   &
1085                   &
1184                   &
1253                   &
1171                   &
1190                   &
--           \\
Test                   & 
\textcolor{cyan}{980}                   &
\textcolor{cyan}{1135}                  &
\textcolor{cyan}{1032}                  &
\textcolor{cyan}{1010}                   &
\textcolor{cyan}{982}                   &
\textcolor{cyan}{892}                   &
\textcolor{cyan}{958}                   &
\textcolor{cyan}{1028}                   &
\textcolor{cyan}{947}                   &
\textcolor{cyan}{1009}                   &
--           \\
\midrule[1.5pt]
~~~~~             &
\multicolumn{11}{c}{MNIST-M (Target domain)}                           \\
\midrule[1.5pt]
Train                    & 
1087                   &
1087                 &
1087                  &
1087                   &
1087                   &
--                   &
--                   &
--                   &
--                   &
--                   &
19039           \\
Validation                    & 
1185                   &
1349                    &
1192                  &
1227                   &
1169                   &
--                   &
--                   &
--                   &
--                   &
--                   &
--           \\
Test                   & 
\textcolor{green}{980}                   &
\textcolor{green}{1135}                  &
\textcolor{green}{1032}                  &
\textcolor{green}{1010}                   &
\textcolor{green}{982}                   &
\textcolor{red}{892}                   &
\textcolor{red}{958}                   &
\textcolor{red}{1028}                   &
\textcolor{red}{947}                   &
\textcolor{red}{1009}                  &
--           \\
\bottomrule
\bottomrule
\end{tabular}
\end{table*}

\begin{table*}[htb]
\centering
\caption{Data split of fetal US standard plane classification (\textbf{anatomies vs. artifacts}). \textcolor{cyan}{Blue} test data is $T_{Source}$, \textcolor{green}{Green} test data is $T_{Target}$ and \textcolor{red}{red} test data is $T_{Target}^{New}$. In this case, all labeled data is $\sim30\%$ of all training data. Train:validation is about 8:2.}
\label{USSplitSA}
\begin{tabular}{ccccccccc}
\toprule
\toprule
~~~~~             &
\multicolumn{7}{c}{Shadow-free fetal US (Source domain)}                           \\
\midrule[1.5pt]
~~~~~                    & 
4CH                   &
Abdominal                   &
LVOT                  &
RVOT                   &
Lips                   &
Femur                   &
Unlabeled           \\
Train                    & 
202                   &
101                 &
101                  &
101                   &
202                  &
202                   &
2125           \\
Validation                    & 
139                   &
63                    &
115                  &
106                   &
166                   &
167                   &
--           \\
Test                   & 
\textcolor{cyan}{50}                   &
\textcolor{cyan}{50}                  &
\textcolor{cyan}{50}                  &
\textcolor{cyan}{50}                   &
\textcolor{cyan}{50}                   &
\textcolor{cyan}{50}                   &
--           \\
\midrule[1.5pt]
~~~~~             &
\multicolumn{7}{c}{Shadow-containing fetal US (Target domain)}                           \\
\midrule[1.5pt]
Train                    & 
--                   &
101                 &
101                  &
101                   &
--                   &
--                   &
710           \\
Validation                    & 
--                   &
119                    &
73                  &
60                   &
--                   &
--                   &
--           \\
Test                   & 
\textcolor{red}{50}                   &
\textcolor{green}{50}                  &
\textcolor{green}{50}                  &
\textcolor{green}{50}                   &
\textcolor{red}{50}                   &
\textcolor{red}{50}                   &
--           \\
\bottomrule
\bottomrule
\end{tabular}
\end{table*}

\begin{table*}[htb]
\centering
\caption{Data split of fetal US standard plane classification (\textbf{anatomies vs. imaging devices}). \textcolor{cyan}{Blue} test data is $T_{Source}$, \textcolor{green}{Green} test data is $T_{Target}$ and \textcolor{red}{red} test data is $T_{Target}^{New}$. In this case, all labeled data is $\sim30\%$ of all training data. Train:validation is about 8:2.}
\label{USSplitCrossdivece}
\begin{tabular}{ccccccc}
\toprule
\toprule
~~~~~             &
\multicolumn{5}{c}{Fetal US from imaging device A (GE Voluson E8)(Source domain)}                           \\
\midrule[1.5pt]
~~~~~                    & 
Abdominal                   &
Brain                  &
Femur                   &
Lips                   &
Unlabeled           \\
Train                    & 
237                   &
237                 &
475                  &
475                   &
3456           \\
Validation                    & 
180                   &
180                    &
420                  &
440                   &
--           \\
Test                   & 
\textcolor{cyan}{100}                   &
\textcolor{cyan}{100}                  &
\textcolor{cyan}{100}                  &
\textcolor{cyan}{100}                   &
--           \\
\midrule[1.5pt]
~~~~~             &
\multicolumn{5}{c}{Fetal US from imaging device B ((Philips EPIQ V7 G)(Target domain)}                           \\
\midrule[1.5pt]
Train                    & 
237                   &
237                 &
--                  &
--                   &
992           \\
Validation                    & 
182                   &
184                    &
--                  &
--                   &
--           \\
Test                   & 
\textcolor{green}{100}                   &
\textcolor{green}{100}                  &
\textcolor{red}{100}                  &
\textcolor{red}{100}                   &
--           \\
\bottomrule
\bottomrule
\end{tabular}
\end{table*}

\section{Semi-/fully-supervised setting in fetal US classification}

\begin{figure*}[htb]
 \centering
 \begin{tabular}{@{\hspace{-1\tabcolsep}}c@{\hspace{0.3\tabcolsep}}c@{\hspace{0.3\tabcolsep}}c@{\hspace{1\tabcolsep}}c@{\hspace{0.3\tabcolsep}}c@{\hspace{0.3\tabcolsep}}c@{\hspace{-1\tabcolsep}}}
 ~~~~~  &
 \raisebox{0.2\height}{\rotatebox[origin=c]{0}{\makecell{~\scalebox{1}{\textbf{Semi-supervised}}}}}  &
 \raisebox{0.2\height}{\rotatebox[origin=c]{0}{\makecell{~\scalebox{1}{\textbf{Fully-supervised}}}}}  &
 ~~~~~                          &
 \raisebox{0.2\height}{\rotatebox[origin=c]{0}{\makecell{~\scalebox{1}{\textbf{Semi-supervised}}}}}  &
 \raisebox{0.2\height}{\rotatebox[origin=c]{0}{\makecell{~\scalebox{1}{\textbf{Fully-supervised}}}}}  \\
 \raisebox{2.6\height}{\rotatebox[origin=c]{90}{\makecell{~\scalebox{1}{$\mathbf{T_{Source}}$}}}} &
  \includegraphics[height=4.2cm, trim=2cm 0cm 3cm 0cm, clip]{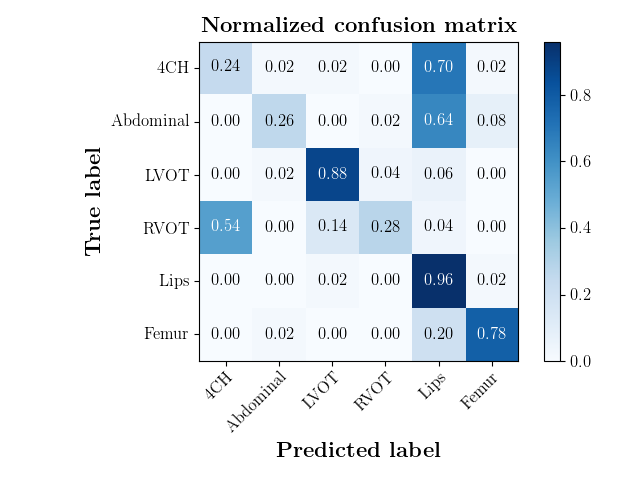} &
  \includegraphics[height=4.2cm, trim=2cm 0cm 1cm 0cm, clip]{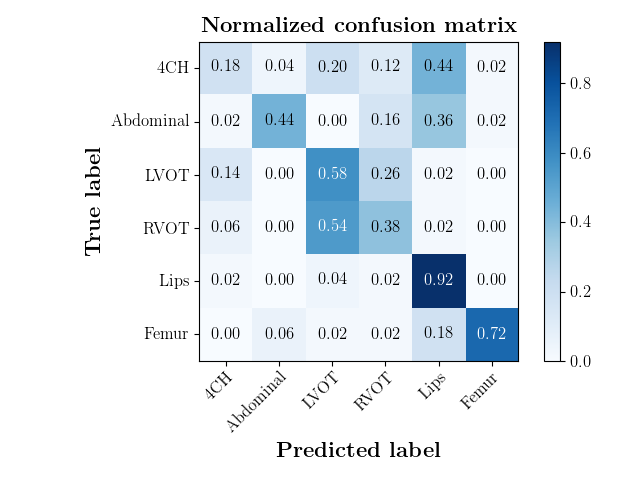} &
  \raisebox{1.2\height}{\rotatebox[origin=c]{90}{\makecell{~\scalebox{1}{$\mathbf{T_{Target}}$ $\&$ $\mathbf{T_{Target}^{New}}$}}}} &
  \includegraphics[height=4.2cm, trim=2cm 0cm 3cm 0cm, clip]{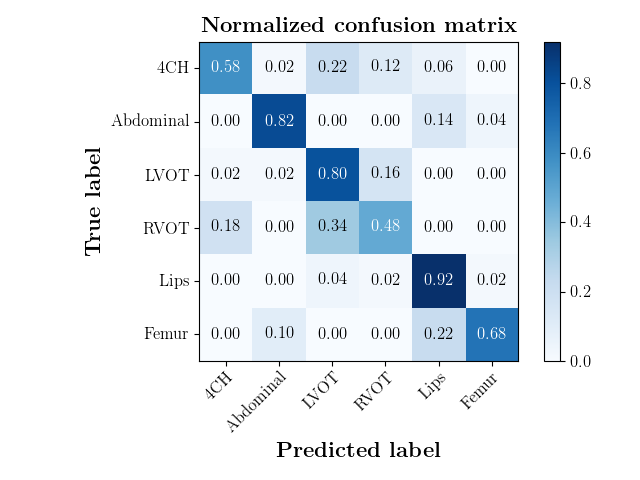} &
  \includegraphics[height=4.2cm, trim=2cm 0cm 1cm 0cm, clip]{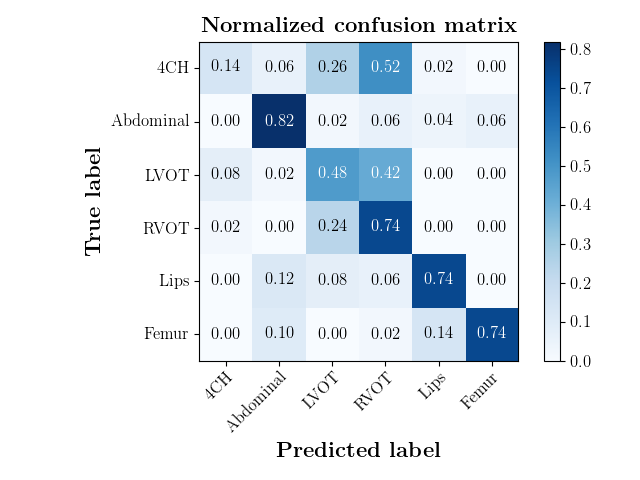}
  \end{tabular}
  \caption{Confusion matrices for fetal US standard plane classification: semi-supervised setting versus fully-supervised setting with MIDNet-VIII model as backbone. For the semi-supervised setting, $30\%$ of training data are labeled data and the rest are unlabeled data. In this experiment the fully-supervised setting only uses $30\%$ labeled data for training, without using unlabeled data.}
  \label{US_CM}
\end{figure*}

We compare the performance of MIDNet-VIII in a semi-supervised setting and a fully-supervised setting for fetal US standard plane classification (separating anatomical features from shadow artifacts features). We observe from Fig.~\ref{US_CM} that the semi-supervised setting yields a better confusion matrix compared with the fully-supervised setting, especially on $T_{Target}$ and $T_{Target}^{New}$. This demonstrates the effectiveness of unlabeled data in our proposed method. One interesting observation is that the classification performance of shadow-containing standard planes are better than that of the corresponding shadow-free standard planes, such as 4CH, Abdominal and RVOT. As we discuss in Sec.\ref{discuss}, this also indicates that shadow artifacts can be more informative than the real anatomies for some categories.

\section{Details of cross-device experiment}
\begin{figure*}[htb]
 \centering
 \setcounter{subfigure}{0}
 \begin{tabular}{@{\hspace{-1\tabcolsep}}c@{\hspace{0.3\tabcolsep}}c@{\hspace{0.7\tabcolsep}}c@{\hspace{0.7\tabcolsep}}c@{\hspace{0.7\tabcolsep}}c@{\hspace{-1\tabcolsep}}}
  \raisebox{1.5\height}{\rotatebox[origin=c]{90}{\makecell{~\scalebox{1}{Device A}}}} &
  \includegraphics[height=2.6cm]{Abdominal_1_2.png} &
  \includegraphics[height=2.6cm]{Brain_1_1.png} &
  \includegraphics[height=2.6cm]{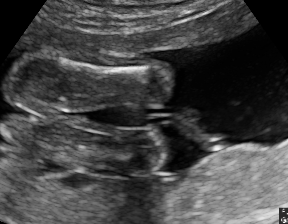} &
  \includegraphics[height=2.6cm]{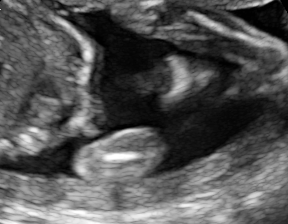} \\
  \raisebox{1.5\height}{\rotatebox[origin=c]{90}{\makecell{~\scalebox{1}{Device B}}}} &
  \stackunder{\includegraphics[height=2.6cm]{Abdominal_2_1_crop.png}}{~\scalebox{1}{Abdominal}} &
  \stackunder{\includegraphics[height=2.6cm]{Brain_2_0_crop.png}}{~\scalebox{1}{Brain}} &
  \stackunder{\includegraphics[height=2.6cm]{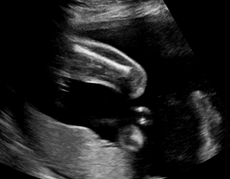}}{~\scalebox{1}{Femur}} &
  \stackunder{\includegraphics[height=2.6cm]{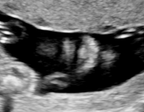}}{~\scalebox{1}{Lips}} 
  \end{tabular}
  \caption{Examples of fetal US dataset for the cross-device experiment. For each standard plane, images are acquired by scanning different patients using different imaging devices. The top row images are random samples of selected standard planes acquired from device A, GE Voluson E8, and the bottom row images are obtained by device B, Philips EPIQ V7 G. These examples show that, except differences in latent features such as noise patterns, different imaging devices acquire images with distinctly different image qualities. For example, device B generate images with more detailed anatomical structures than device A. Note that the shown examples are from different patients and not paired during training. 
  }
  \label{DataPresentCrossDevice}
\end{figure*}

In this section, we present the dataset that is used in the cross-device experiment and show more detailed results. We utilize fetal US standard planes consisting of $~\sim6k$ images, acquired by two different imaging devices, device A is a GE Voluson E8 and device B is a Philips EPIQ V7 G. We show some image examples from randomly selected patients in Fig.~\ref{DataPresentCrossDevice}. The data split is shown in Table.~\ref{USSplitCrossdivece}.

\begin{figure*}[htb]
 \centering
 \setcounter{subfigure}{0}
 \subfloat[Test on $\mathbf{T_{Source}}$]{
 \begin{tabular}{@{\hspace{-1\tabcolsep}}c@{\hspace{0.3\tabcolsep}}c@{\hspace{-1\tabcolsep}}}
  \stackunder{\includegraphics[height=3.5cm, trim=2cm 0cm 0cm 0cm, clip]{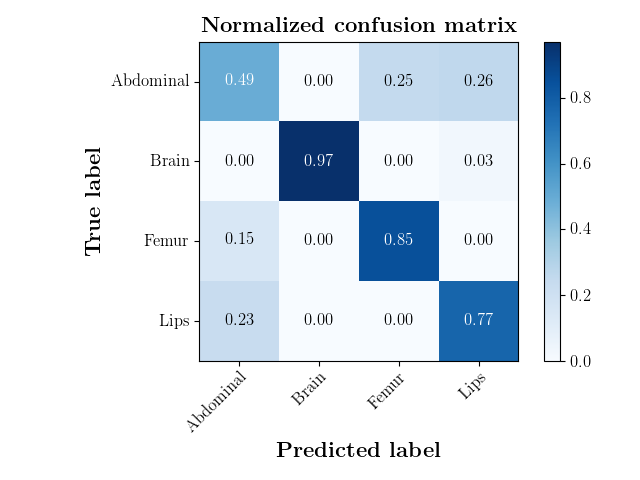}}{~\scalebox{1}{\textit{Source only}}} &
  \stackunder{\includegraphics[height=3.5cm, trim=2cm 0cm 0cm 0cm, clip]{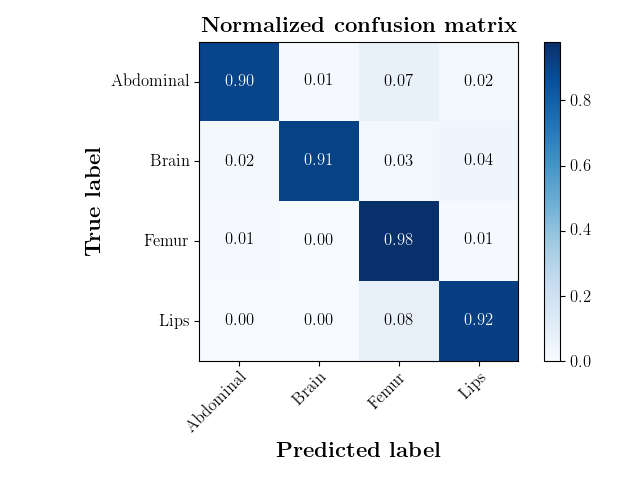}}{~\scalebox{1}{MIDNet-VIII}} 
  \end{tabular}
  } 
  \setcounter{subfigure}{1}
  \subfloat[Test on $\mathbf{T_{Target}}$ $\&$ $\mathbf{T_{Target}^{New}}$]{
 \begin{tabular}{@{\hspace{-1\tabcolsep}}c@{\hspace{0.3\tabcolsep}}c@{\hspace{0.3\tabcolsep}}}
  \stackunder{\includegraphics[height=3.5cm, trim=2cm 0cm 0cm 0cm, clip]{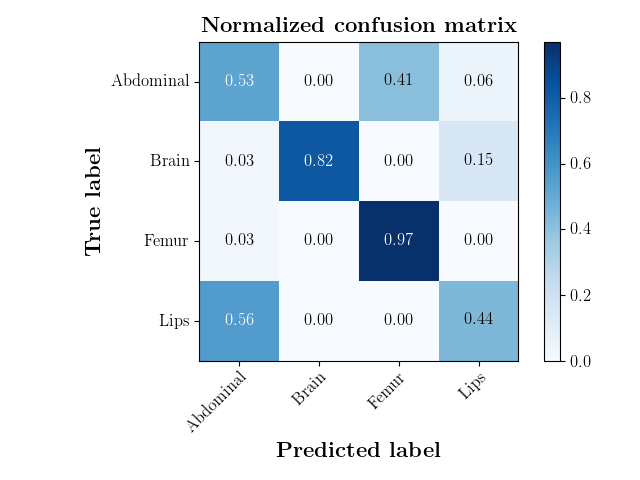}}{~\scalebox{1}{\textit{Source only}}} &
  \stackunder{\includegraphics[height=3.5cm, trim=2cm 0cm 0cm 0cm, clip]{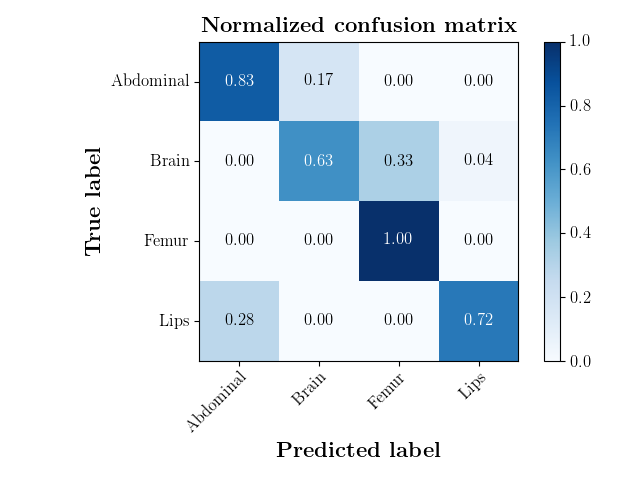}}{~\scalebox{1}{MIDNet-VIII}} 
  \end{tabular}
  }
  \caption{Confusion matrices for the cross-device experiment: \textit{Source only} versus MIDNet-VIII. For MIDNet-VIII, $30\%$ of the training data is labeled data and the rest is unlabeled. \textit{Source only} in this experiment uses the $30\%$ labeled data for training, without using unlabeled data.
  }
  \label{iFind12_CM}
\end{figure*}

We verify the effectiveness of our method by comparing the MIDNet-VIII model with the baseline \textit{Source only}. \textit{Source only} is utilized to demonstrate that a domain shift problem exists between the two different imaging devices. We train the MIDNet-VIII model using a semi-supervised setting where $30\%$ of the training data is labeled and the rest is unlabeled. MIDNet-VIII is optimized for 200 training epoch. We have shown the average accuracy in the main paper (Table.~\ref{iFind12_table}). Here, we further show the confusion matrices of the three groups of test data in Fig.~\ref{iFind12_CM}. From this figure, we observe that the classification performance is improved for most of the classes in MIDNet-VIII on all groups of test data. One interesting observation is that the classification performance of \textit{Source only} is slightly better than that of MIDNet-VIII for the Brain category. This can be observed in both domains which indicates that, on this dataset, MIDNet-VIII may trade the classification performance for the Brain category for improvements of the other categories.   

\section{Additional classification results}

\begin{figure*}[htb]
 \centering
 \setcounter{subfigure}{0}
 \subfloat[Handwritten digits $\mathbf{T_{Target}^{New}}$ (target domain is MNIST-M)]{
 \begin{tabular}{@{\hspace{-1\tabcolsep}}c@{\hspace{0.3\tabcolsep}}c@{\hspace{0.3\tabcolsep}}c@{\hspace{0.3\tabcolsep}}c@{\hspace{0.3\tabcolsep}}c@{\hspace{0.3\tabcolsep}}c@{\hspace{0.3\tabcolsep}}c@{\hspace{0.3\tabcolsep}}c@{\hspace{0.3\tabcolsep}}c@{\hspace{0.3\tabcolsep}}c@{\hspace{0.3\tabcolsep}}c@{\hspace{-1\tabcolsep}}}
 \raisebox{0.2\height}{\rotatebox[origin=c]{0}{\makecell{~\scalebox{1}{\textbf{Ground truth}}}}}  &
 \raisebox{0.2\height}{\rotatebox[origin=c]{0}{\makecell{~\scalebox{1.2}{\textbf{5}}}}}  &
 \raisebox{0.2\height}{\rotatebox[origin=c]{0}{\makecell{~\scalebox{1.2}{\textbf{5}}}}}  &
 \raisebox{0.2\height}{\rotatebox[origin=c]{0}{\makecell{~\scalebox{1.2}{\textbf{6}}}}}   &
 \raisebox{0.2\height}{\rotatebox[origin=c]{0}{\makecell{~\scalebox{1.2}{\textbf{6}}}}}  &
 \raisebox{0.2\height}{\rotatebox[origin=c]{0}{\makecell{~\scalebox{1.2}{\textbf{7}}}}}  &
 \raisebox{0.2\height}{\rotatebox[origin=c]{0}{\makecell{~\scalebox{1.2}{\textbf{7}}}}}   &
 \raisebox{0.2\height}{\rotatebox[origin=c]{0}{\makecell{~\scalebox{1.2}{\textbf{8}}}}}  &
 \raisebox{0.2\height}{\rotatebox[origin=c]{0}{\makecell{~\scalebox{1.2}{\textbf{8}}}}}  &
 \raisebox{0.2\height}{\rotatebox[origin=c]{0}{\makecell{~\scalebox{1.2}{\textbf{9}}}}}   &
 \raisebox{0.2\height}{\rotatebox[origin=c]{0}{\makecell{~\scalebox{1.2}{\textbf{9}}}}}  \\
 \raisebox{1.5\height}{\rotatebox[origin=c]{0}{\makecell{~\scalebox{1}{\textbf{True positive}}}}}   &
  \stackunder{\includegraphics[height=1.5cm]{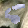}}{~\scalebox{1}{5}} &
  \stackunder{\includegraphics[height=1.5cm]{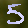}}{~\scalebox{1}{5}} &
  \stackunder{\includegraphics[height=1.5cm]{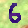}}{~\scalebox{1}{6}} &
  \stackunder{\includegraphics[height=1.5cm]{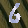}}{~\scalebox{1}{6}} &
  \stackunder{\includegraphics[height=1.5cm]{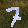}}{~\scalebox{1}{7}} &
  \stackunder{\includegraphics[height=1.5cm]{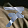}}{~\scalebox{1}{7}} &
  \stackunder{\includegraphics[height=1.5cm]{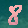}}{~\scalebox{1}{8}} &
  \stackunder{\includegraphics[height=1.5cm]{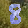}}{~\scalebox{1}{8}} &
  \stackunder{\includegraphics[height=1.5cm]{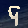}}{~\scalebox{1}{9}} &
  \stackunder{\includegraphics[height=1.5cm]{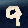}}{~\scalebox{1}{9}} \\
  \raisebox{1.5\height}{\rotatebox[origin=c]{0}{\makecell{~\scalebox{1}{\textbf{False positive}}}}}   &
  \stackunder{\includegraphics[height=1.5cm]{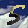}}{~\scalebox{1}{0}} &
  \stackunder{\includegraphics[height=1.5cm]{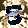}}{~\scalebox{1}{3}} &
  \stackunder{\includegraphics[height=1.5cm]{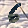}}{~\scalebox{1}{4}} &
  \stackunder{\includegraphics[height=1.5cm]{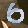}}{~\scalebox{1}{0}} &
  \stackunder{\includegraphics[height=1.5cm]{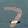}}{~\scalebox{1}{2}} &
  \stackunder{\includegraphics[height=1.5cm]{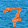}}{~\scalebox{1}{1}} &
  \stackunder{\includegraphics[height=1.5cm]{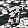}}{~\scalebox{1}{3}} &
  \stackunder{\includegraphics[height=1.5cm]{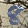}}{~\scalebox{1}{4}} &
  \stackunder{\includegraphics[height=1.5cm]{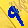}}{~\scalebox{1}{4}} &
  \stackunder{\includegraphics[height=1.5cm]{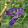}}{~\scalebox{1}{4}} 
  \end{tabular}
  } 
  \setcounter{subfigure}{1}
  \subfloat[Fetal US data $\mathbf{T_{Target}^{New}}$ (target domain is shadow-containing image)]{
 \begin{tabular}{@{\hspace{-1\tabcolsep}}c@{\hspace{0.3\tabcolsep}}c@{\hspace{0.3\tabcolsep}}c@{\hspace{0.3\tabcolsep}}c@{\hspace{0.3\tabcolsep}}c@{\hspace{0.3\tabcolsep}}c@{\hspace{0.3\tabcolsep}}c@{\hspace{0.3\tabcolsep}}}
 \raisebox{0.2\height}{\rotatebox[origin=c]{0}{\makecell{~\scalebox{1}{\textbf{Ground truth}}}}}  &
 \raisebox{0.2\height}{\rotatebox[origin=c]{0}{\makecell{~\scalebox{1.2}{\textbf{4CH}}}}}  &
 \raisebox{0.2\height}{\rotatebox[origin=c]{0}{\makecell{~\scalebox{1.2}{\textbf{4CH}}}}}  &
 \raisebox{0.2\height}{\rotatebox[origin=c]{0}{\makecell{~\scalebox{1.2}{\textbf{Femur}}}}}   &
 \raisebox{0.2\height}{\rotatebox[origin=c]{0}{\makecell{~\scalebox{1.2}{\textbf{Femur}}}}}  &
 \raisebox{0.2\height}{\rotatebox[origin=c]{0}{\makecell{~\scalebox{1.2}{\textbf{Lips}}}}}  &
 \raisebox{0.2\height}{\rotatebox[origin=c]{0}{\makecell{~\scalebox{1.2}{\textbf{Lips}}}}}   \\
 \raisebox{2.5\height}{\rotatebox[origin=c]{0}{\makecell{~\scalebox{1}{\textbf{True positive}}}}}   &
  \stackunder{\includegraphics[height=2cm]{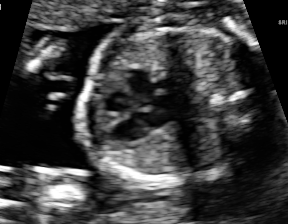}}{~\scalebox{1}{4CH}} &
  \stackunder{\includegraphics[height=2cm]{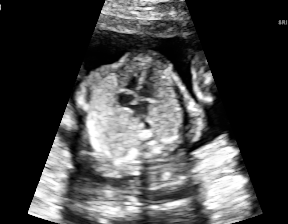}}{~\scalebox{1}{4CH}} &
  \stackunder{\includegraphics[height=2cm]{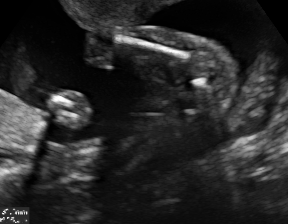}}{~\scalebox{1}{Femur}} &
  \stackunder{\includegraphics[height=2cm]{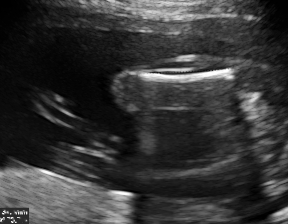}}{~\scalebox{1}{Femur}} &
  \stackunder{\includegraphics[height=2cm]{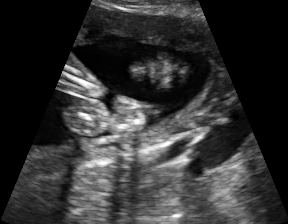}}{~\scalebox{1}{Lips}} &
  \stackunder{\includegraphics[height=2cm]{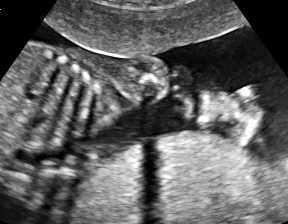}}{~\scalebox{1}{Lips}} \\
  \raisebox{2.5\height}{\rotatebox[origin=c]{0}{\makecell{~\scalebox{1}{\textbf{False positive}}}}}   &
  \stackunder{\includegraphics[height=2cm]{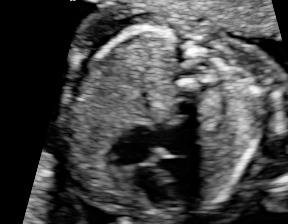}}{~\scalebox{1}{Abdominal}} &
  \stackunder{\includegraphics[height=2cm]{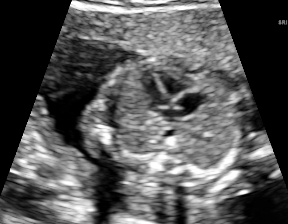}}{~\scalebox{1}{LVOT}} &
  \stackunder{\includegraphics[height=2cm]{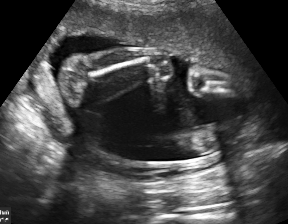}}{~\scalebox{1}{Abdominal}} &
  \stackunder{\includegraphics[height=2cm]{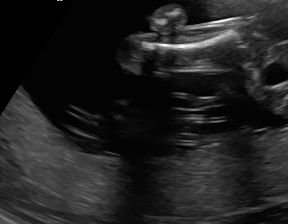}}{~\scalebox{1}{Lips}} &
  \stackunder{\includegraphics[height=2cm]{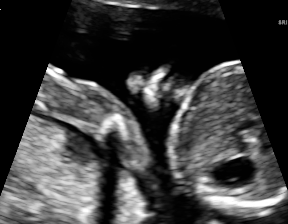}}{~\scalebox{1}{LVOT}} &
  \stackunder{\includegraphics[height=2cm]{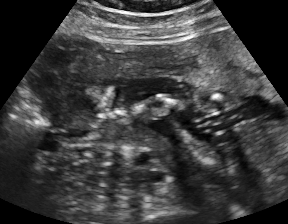}}{~\scalebox{1}{Femur}} 
  \end{tabular}
  }
  \setcounter{subfigure}{2}
  \subfloat[Fetal US data $\mathbf{T_{Target}^{New}}$ (target domain is image from device B, Philips EPIQ V7 G)]{
 \begin{tabular}{@{\hspace{-1\tabcolsep}}c@{\hspace{0.3\tabcolsep}}c@{\hspace{0.3\tabcolsep}}c@{\hspace{0.3\tabcolsep}}c@{\hspace{-1\tabcolsep}}}
 \raisebox{0.2\height}{\rotatebox[origin=c]{0}{\makecell{~\scalebox{1}{\textbf{Ground truth}}}}}  &
 \raisebox{0.2\height}{\rotatebox[origin=c]{0}{\makecell{~\scalebox{1.2}{\textbf{Femur}}}}}  &
 \raisebox{0.2\height}{\rotatebox[origin=c]{0}{\makecell{~\scalebox{1.2}{\textbf{Lips}}}}}   &
 \raisebox{0.2\height}{\rotatebox[origin=c]{0}{\makecell{~\scalebox{1.2}{\textbf{Lips}}}}}   \\
 \raisebox{2.5\height}{\rotatebox[origin=c]{0}{\makecell{~\scalebox{1}{\textbf{True positive}}}}}   &
  \stackunder{\includegraphics[height=2cm]{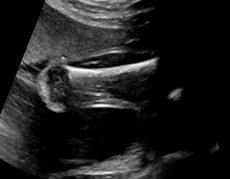}}{~\scalebox{1}{Femur}} &
  \stackunder{\includegraphics[height=2cm]{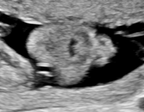}}{~\scalebox{1}{Lips}} &
  \stackunder{\includegraphics[height=2cm]{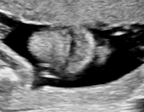}}{~\scalebox{1}{Lips}}  \\
  \raisebox{2.5\height}{\rotatebox[origin=c]{0}{\makecell{~\scalebox{1}{\textbf{False positive}}}}}   &
  \stackunder{\includegraphics[height=2cm]{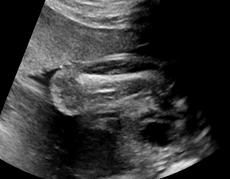}}{~\scalebox{1}{Abdominal}} &
  \stackunder{\includegraphics[height=2cm]{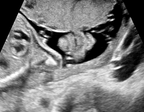}}{~\scalebox{1}{Abdominal}} &
  \stackunder{\includegraphics[height=2cm]{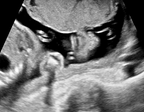}}{~\scalebox{1}{Abdominal}} 
  \end{tabular}
  }
  \caption{Inference examples of three classification tasks on two datasets using MIDNet-VIII. We show true positive and false positive samples from $T_{Target}^{New}$. \textbf{Ground truth} contains ground truth labels and labels under images are predicted labels. (a) Handwritten digits classification. (b) Fetal US standard plane classification that separate anatomical features from shadow artifacts features. (c) Fetal US standard plane classification that disjoin anatomical featurs from imaging device features. 
  }
  \label{TPFP}
\end{figure*}

We further show randomly selected true positive and false positive inference examples (images) from the three classification tasks in Fig.~\ref{TPFP}.


\end{document}